%% file: main.tex
\documentclass[10pt,twocolumn,letterpaper]{article}

\usepackage{cvpr}      

\usepackage{colortbl}
\usepackage{graphicx}
\usepackage{amsmath,amssymb}
\usepackage{booktabs}
\usepackage{multirow}
\usepackage{xcolor}
\usepackage{subcaption}
\usepackage{makecell}
\usepackage{tabularx}
\usepackage{array}
\usepackage{xltabular}
\usepackage{algorithm}
\usepackage{algorithmic}
\usepackage{newfloat}
\usepackage{listings}

\newcolumntype{Y}{>{\centering\arraybackslash}X}

\newcommand{\cmark}{\textcolor{green!60!black}{$\checkmark$}}
\newcommand{\xmark}{\textcolor{red!80!black}{$\times$}}

\newcommand{\ours}{\textit{PersonaShot}}
\newcommand{\autoeval}{PersonaShot-AutoEval}

\definecolor{cvprblue}{rgb}{0.21,0.49,0.74}
\usepackage[pagebackref,breaklinks,colorlinks,allcolors=cvprblue]{hyperref}

\def\paperID{****} 
\def\confName{CVPR}
\def\confYear{2026}

\title{PersonaShot: Benchmarking Person-Centric Narrative Continuity in \\Multi-Shot Video Generation}

\author{
    Yuji Wang\textsuperscript{1}\thanks{Equal contribution.} \quad
    Yuheng Chen\textsuperscript{1}\footnotemark[1] \quad
    Teng Hu\textsuperscript{1} \quad
    Ran Yi\textsuperscript{1}\thanks{Project lead.} \quad
    Yijia Hong\textsuperscript{1} \quad
    Han Feng\textsuperscript{2} \\
    Weijian Cao\textsuperscript{2} \quad
    Chengjie Wang\textsuperscript{2} \quad
    Lizhuang Ma\textsuperscript{1}\thanks{Corresponding author.} \quad
    Jiangning Zhang\textsuperscript{3} \\
    \\
    \textsuperscript{1}Shanghai Jiao Tong University \quad
    \textsuperscript{2}Tencent Youtu Lab \quad
    \textsuperscript{3}Zhejiang University \\
    {\tt\small \{yujiwang@sjtu.edu.cn, ma-lz@cs.sjtu.edu.cn\}} 
}

\begin{document}
\maketitle

\input{Secs/00_abstract}
\input{Secs/01_introduction}
\input{Secs/02_related_work}
\input{Secs/03_dataset}
\input{Secs/04_methodology}
\input{Secs/05_experiments}
\input{Secs/06_conclusion}

{
    \small
    \bibliographystyle{ieeenat_fullname}
    \bibliography{main}
}

\clearpage
\appendix
\noindent\textbf{\Large Supplementary Material}
\vspace{10pt}

\input{Suppl_secs/detailed_related_work}
\input{Suppl_secs/implementation_detail}
\input{Suppl_secs/detailed_metrics}
\input{Suppl_secs/extended_exp}
\input{Suppl_secs/human_study}

\end{document}

%% file: Secs/00_abstract.tex
\begin{abstract}
Video generation is rapidly evolving from single-shot clips to multi-shot narratives, where the human character serves as the core narrative anchor. However, existing benchmarks mainly assess character appearance or individual-shot quality, without measuring whether physical and emotional states remain coherent across cuts. They also rarely provide criterion-specific evaluation methods, although physical continuity, facial dynamics, and cinematic relations require different visual, temporal, and relational evidence. To address these limitations, we introduce \textbf{\ours{}}, the first person-centric benchmark for narrative continuity in multi-shot video generation. \ours{} contains approximately 1,000 multi-shot segments and 16 metrics spanning physical continuity, affective dynamics, and cinematic grammar. 
\textbf{\textit{1)} Narrative Continuity Benchmark:} We evaluate character coherence across three temporal levels: within-shot states, cross-shot transitions, and sequence-level trajectories. 
\textbf{\textit{2)} Human-Aligned Specialist Evaluators:} We distill reasoning from a large multimodal teacher into lightweight criterion-specific evaluators, each grounded in the visual, temporal, or relational evidence required by its metric, and align them with expert human judgments. 
\textbf{\textit{3)} Systematic Evaluation and Insights:} Our evaluation reveals distinct capability profiles across state-of-the-art models and a clear gap between perceptual quality and cross-shot narrative continuity. Even visually compelling videos frequently exhibit physical-state resets, abrupt affective shifts, and broken cinematic relations across shots. Human studies further demonstrate strong agreement between our evaluators and expert judgments.
\vspace{-1mm}
\noindent \textbf{Project Page:} \url{https://rain152.github.io/PersonaShot/}
\end{abstract}

%% file: Secs/01_introduction.tex
\section{Introduction}
\label{sec:intro}

\begin{figure}[t]
    \centering
    \includegraphics[width=\linewidth]{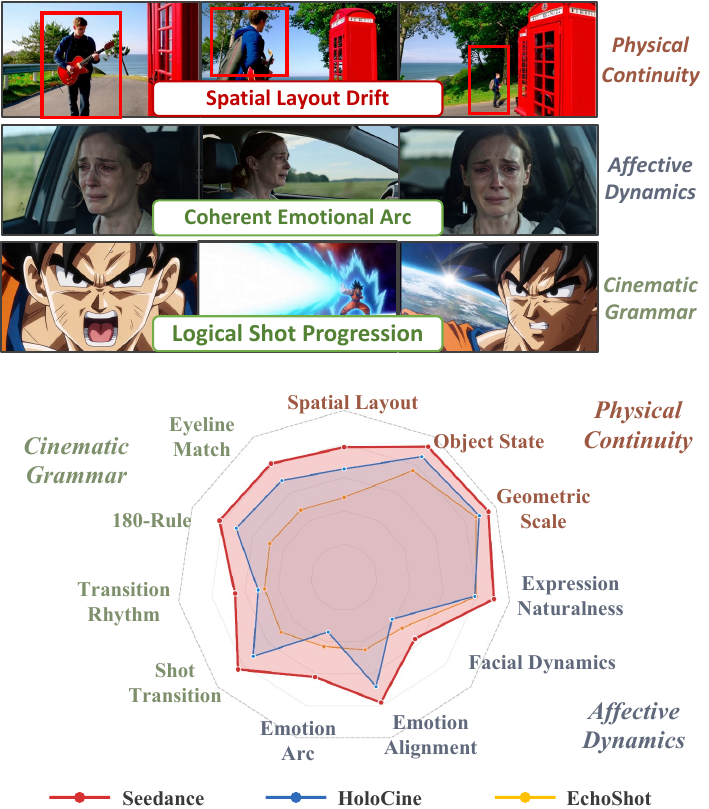}
    \caption{\textbf{Overview of PersonaShot's person-centric evaluation.} 
    \textbf{Top:} Qualitative examples illustrating our three core dimensions: physical continuity, affective dynamics, and cinematic grammar. 
    \textbf{Bottom:} Quantitative comparison across fine-grained metrics for specific state-of-the-arts, revealing their distinct capability profiles.}
    \label{fig:teaser}
    \vspace{-4mm}
\end{figure}

\begin{table*}[t]
\centering
\resizebox{\linewidth}{!}{%
\begingroup
\renewcommand{\arraystretch}{1.1} 
\setlength{\aboverulesep}{0pt}
\setlength{\belowrulesep}{0pt}
\setlength{\cmidrulesep}{0.15ex}
\begin{tabular}{@{} l cccc c cc cc c @{}}
\toprule
\multirow{3}{*}{\textbf{Benchmark}}
& \multicolumn{4}{c}{\textbf{Input Modalities}}
& \multicolumn{6}{c}{\textbf{Evaluation Capabilities}} \\
\cmidrule(lr){2-5}\cmidrule(l){6-11}
& \multirow{2}{*}{\thead{Ref.\\Image}}
& \multirow{2}{*}{\thead{Ref.\\Audio}}
& \multirow{2}{*}{\thead{First-\\Frame}}
& \multirow{2}{*}{\thead{Multi-\\Shot}}
& \multirow{2}{*}{\thead{Person-\\Centric}}
& \multicolumn{2}{c}{\textbf{Physics}}
& \multicolumn{2}{c}{\textbf{Emotion}}
& \multirow{2}{*}{\thead{Cinematic\\Grammar}} \\
\cmidrule(lr){7-8}\cmidrule(lr){9-10}
& & & & & & Intra & Cross & Coarse & AU-Aware & \\
\midrule
VBench~\cite{huang2024vbench}
& \xmark & \xmark & \xmark & \xmark & \xmark
& \xmark & \xmark & \xmark & \xmark & \xmark \\
HumanVBench~\cite{zhou2024humanvbench}
& \cmark & \xmark & \xmark & \xmark & \cmark
& \xmark & \xmark & \cmark & \xmark & \xmark \\
HVEval~\cite{wu2025hveval}
& \xmark & \xmark & \xmark & \xmark & \cmark
& \xmark & \xmark & \cmark & \xmark & \xmark \\
VideoPhy-2~\cite{bansal2025videophy}
& \cmark & \xmark & \xmark & \xmark & \xmark
& \cmark & \xmark & \xmark & \xmark & \xmark \\
MSVBench~\cite{shi2026msvbench}
& \cmark & \xmark & \xmark & \cmark & \xmark
& \xmark & \xmark & \xmark & \xmark & \xmark \\
MSAVBench~\cite{wei2026msavbench}
& \cmark & \cmark & \xmark & \cmark & \xmark
& \xmark & \xmark & \cmark & \xmark & \xmark \\
UniVBench~\cite{wei2026univbench}
& \cmark & \xmark & \xmark & \cmark & \xmark
& \xmark & \xmark & \cmark & \xmark & \xmark \\
ViStoryBench~\cite{zhuang2026vistorybench}
& \cmark & \xmark & \xmark & \cmark & \cmark
& \xmark & \xmark & \xmark & \xmark & \xmark \\
MuSS~\cite{zhang2026muss} & \cmark & \xmark & \cmark & \cmark & \xmark & \xmark & \cmark & \xmark & \xmark & \cmark \\
\midrule
\rowcolor{gray!15} 
\textbf{\ours{} (Ours)}
& \textbf{\cmark} & \textbf{\cmark} & \textbf{\cmark}
& \textbf{\cmark} & \textbf{\cmark}
& \textbf{\cmark} & \textbf{\cmark}
& \textbf{\cmark} & \textbf{\cmark} & \textbf{\cmark} \\
\bottomrule
\end{tabular}
\endgroup
}
\caption{\textbf{Comparison of \ours{} with existing video generation benchmarks.}
\textbf{First-Frame} denotes shot-level image conditioning;
\textbf{Intra/Cross} denote within-shot plausibility and cross-shot continuity;
\textbf{AU-Aware} uses facial Action Units, while
\textbf{Cinematic Grammar} covers the 180-degree rule, eyeline matching, and transition logic.}
\label{tab:benchmark_comparison}
\end{table*}

Video generation~\cite{wan2025wan,xing2024survey} is rapidly evolving from single-shot clips to multi-shot narratives~\cite{wu2025cinetrans,wang2026multishotmaster}. In these sequences, the \textbf{human character} serves as the core narrative anchor. A character's physical continuity, emotional progression, and cinematic framing across cuts jointly determine sequence coherence. Inconsistencies in these aspects can break the connection between shots and undermine viewer immersion. 
Yet, no existing benchmark provides a dedicated evaluation of character coherence across shot boundaries. 

Recent benchmarks~\cite{bansal2025videophy,huang2024vbench} cover perceptual quality, motion, physical plausibility, and scene consistency, but remain insufficient for person-centric multi-shot generation. \textit{First,} they mainly assess character appearance or individual-shot quality, without measuring how physical and emotional states evolve across cuts. 
\textit{Second,} physical continuity, facial dynamics, and cinematic relations require different visual, temporal, and relational evidence, yet existing benchmarks rarely provide criterion-specific evaluation methods. 

These limitations motivate the three dimensions illustrated in Fig.~\ref{fig:teaser}-Top: 
\textbf{\textit{1)} Cross-shot physical continuity:} existing physics benchmarks mainly assess within-shot plausibility, without determining whether character position, body state, object interaction, and relative scale remain consistent across cuts. Thus, teleportation and unexplained state changes may remain unpenalized. 
\textbf{\textit{2)} Fine-grained affective dynamics:} existing emotion metrics rely largely on coarse categories, overlooking subtle facial changes, micro-expressions, and continuous emotional progression. 
\textbf{\textit{3)} Cinematic grammar:} multi-shot narratives depend on the 180-degree rule, eyeline matching, screen-direction consistency, and transition logic, which cannot be evaluated by treating shots independently. As summarized in Table~\ref{tab:benchmark_comparison}, existing benchmarks do not jointly cover realistic input modalities and these fine-grained person-centric capabilities.

To bridge these gaps, we introduce \textbf{\ours{}}, a person-centric benchmark for narrative continuity in multi-shot video generation. \ours{} evaluates character coherence across within-shot states, cross-shot transitions, and sequence-level trajectories. It contains approximately 1,000 multi-shot segments and 16 metrics spanning physical continuity, affective dynamics, and cinematic grammar. We further distill structured reasoning from a large multimodal teacher into lightweight specialist evaluators, each grounded in the visual, temporal, or relational evidence required by its criterion. As shown in Fig.~\ref{fig:teaser}-Bottom, current models exhibit distinct capability profiles: even visually compelling outputs may contain physical-state resets, abrupt affective shifts, and broken cinematic relations. These findings demonstrate that perceptual quality alone is insufficient to measure multi-shot narrative continuity. 
In summary, our main contributions are as follows:
\begin{itemize}[leftmargin=*,itemsep=2pt]
\item \textbf{Narrative Continuity Benchmark.} We introduce the first person-centric benchmark for multi-shot video generation, comprising approximately 1,000 segments and 16 metrics across physical, affective, and cinematic dimensions.

\item \textbf{Human-Aligned Specialist Evaluators.} We distill reasoning from a multimodal teacher into lightweight criterion-specific evaluators grounded in the visual, temporal, or relational evidence required by each metric, improving agreement with expert human judgments.

\item \textbf{Systematic Evaluation and Insights.} We reveal distinct capability profiles across state-of-the-art models and a clear gap between perceptual quality and cross-shot narrative continuity. We commit to release the benchmark, evaluators, and code to facilitate future research.

\end{itemize}

%% file: Secs/02_related_work.tex
\section{Related Work}
\label{sec:related}

\begin{figure*}[t]
\centering
\includegraphics[width=\linewidth]{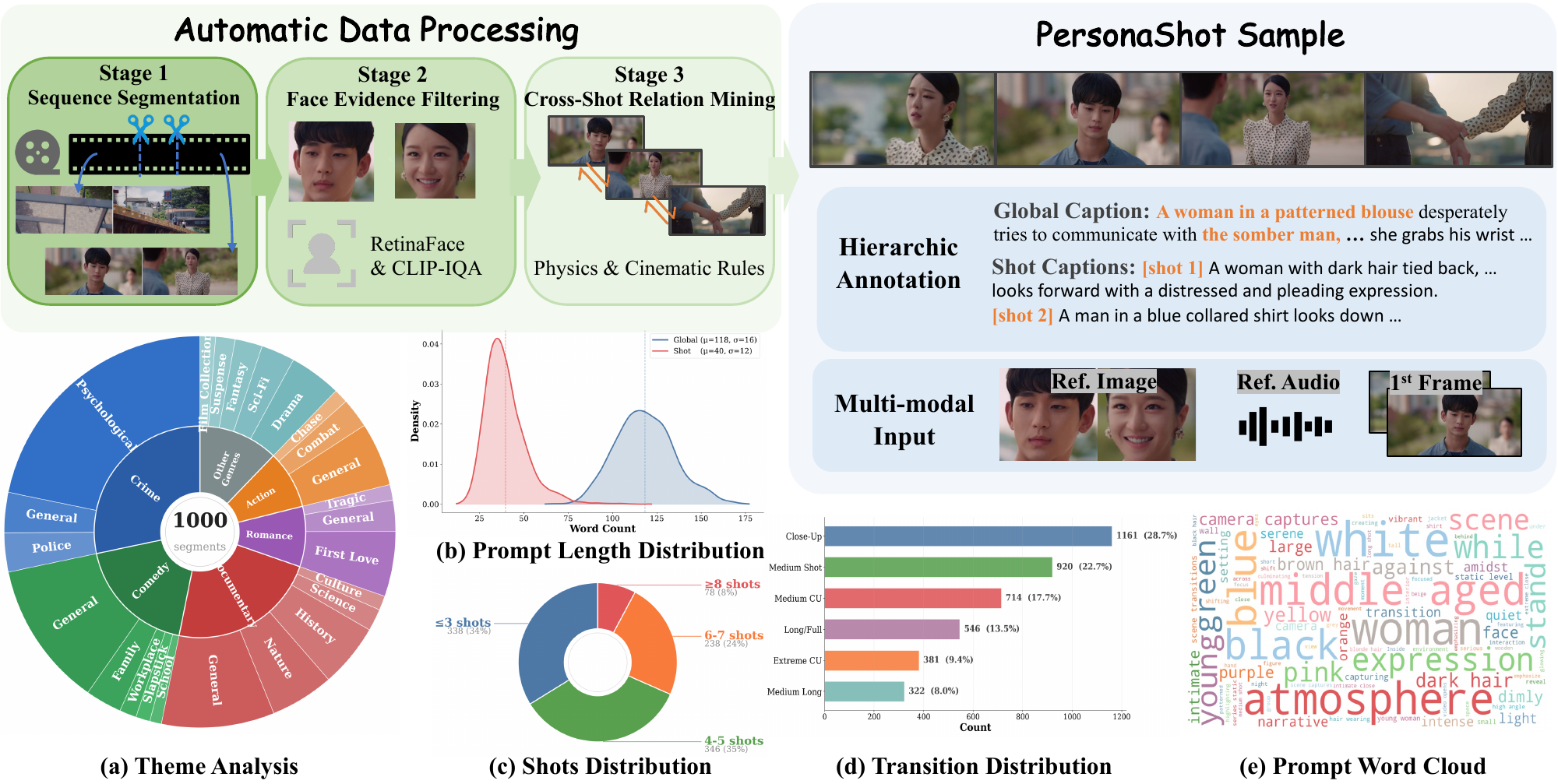}
\caption{\textbf{Overview of the PersonaShot benchmark.} 
\textbf{Top-Left:} Our three-stage data processing pipeline filters multi-shot segments via sequence continuity, face visibility, and cross-shot relations. 
\textbf{Top-Right:} A curated sample featuring hierarchical text annotations and multi-modal conditioning inputs. 
\textbf{Bottom:} Key benchmark statistics.}
\label{fig:data_statistics}
\end{figure*}

\noindent\textbf{Video Generation Benchmarks.}
Early video generation benchmarks, such as VBench~\cite{huang2024vbench} and EvalCrafter~\cite{liu2024evalcrafter}, established multidimensional evaluation protocols for single-shot videos, covering perceptual quality, motion, and semantic alignment. HumanVBench~\cite{zhou2024humanvbench} and HVEval~\cite{wu2025hveval} further introduced human-centric criteria, while recent benchmarks, including MSVBench~\cite{shi2026msvbench}, UniVBench~\cite{wei2026univbench}, and MSAVBench~\cite{wei2026msavbench}, extended evaluation to multi-shot generation and audio-visual consistency. Despite this progress, existing benchmarks remain limited in explicitly modeling how a character's state persists and evolves across shot boundaries. In particular, cross-shot physical continuity, AU-aware affective dynamics, and relational cinematic grammar have not been jointly evaluated within a person-centric framework. \ours{} addresses this gap by assessing character coherence through within-shot states, cross-shot transitions, and sequence-level trajectories.

\noindent\textbf{Multi-Shot Video Generation.}
Recent multi-shot generation systems differ mainly in how narrative and shot structures are specified. Some methods rely on explicit shot-level prompts or interactive controls, including EchoShot~\cite{wang2026echoshot}, ShotStream~\cite{luo2026shotstream}, and LongLive~\cite{yang2025longlive}. Others combine global narratives with per-shot descriptions, such as MultiShotMaster~\cite{wang2026multishotmaster}, HoloCine~\cite{meng2026holocine}, and CineTrans~\cite{wu2025cinetrans}. More automated approaches, including STAGE~\cite{zhang2026stage} and VGoT~\cite{zheng2024vgot}, decompose high-level stories into structured shot sequences. Although these systems offer different balances between user control and automation, they share the challenge of maintaining character coherence across cuts. We evaluate representative methods under a unified protocol to identify their strengths and limitations in physical continuity, affective dynamics, and cinematic grammar.

%% file: Secs/03_dataset.tex
\section{PersonaShot Benchmark Curation}
\label{sec:dataset}

\noindent\textbf{Data Sources.}
Many video generation benchmarks rely on a \emph{prompt-first} strategy, using LLMs to synthesize text prompts for predefined evaluation concepts~\cite{bansal2025videophy}. While highly scalable, text prompts struggle to adequately specify and verify subtle cross-shot relations, such as fine-grained facial dynamics and spatial continuity. 

In contrast, \ours{} adopts a \emph{video-first} paradigm grounded in two large-scale corpora: CineDance-1M~\cite{chen2026cinedance}, which provides professionally edited sequences with rich transition and character annotations, and OpenHumanVid~\cite{li2025openhumanvid}, which expands action and scene diversity. We re-segment these videos, track recurring characters, and rigorously filter samples to guarantee sufficient visual evidence for multi-dimensional evaluation.

\noindent\textbf{Automatic Data Processing.}
We construct reliable multi-shot evaluation samples using a three-stage pipeline. The first stage forms candidate sequences, while subsequent stages verify the presence of requisite evidence for physical, affective, or cinematic evaluation. A segment lacking evidence for one dimension may still be retained for others.

\begin{itemize}[leftmargin=*,itemsep=2pt,topsep=2pt]

\item \textbf{Stage 1: Sequence Segmentation and Character Continuity.}
We verify source shot boundaries using TransNetV2~\cite{soucek2024transnet} and segment videos into sequences of 2--15 shots, ensuring each contains at least one cross-shot transition. To guarantee narrative coherence, we retain only segments where at least one recurring character appears across multiple shots within a coherent local event, excluding disjointed shot collections or abrupt topic shifts.

\item \textbf{Stage 2: Face Visibility and Quality.}
AU-aware affective analysis requires clear facial observations across related shots. We uniformly sample frames from each shot and detect faces using RetinaFace~\cite{deng2020retinaface}, with a confidence threshold of 0.85 and a minimum face area of 5\% of the frame. A segment is retained for affective evaluation when faces satisfying these criteria appear in at least 60\% of its shots and in no fewer than two shots. We further apply CLIP-IQA~\cite{wang2023clipiqa} to remove heavily blurred or degraded face crops. The resulting face boxes, crops, and visibility records support subsequent affective annotation and evaluation.

\item \textbf{Stage 3: Cross-Shot Relation Selection.}
We leverage Qwen-based~\cite{team2026qwen3,bai2023qwen} multimodal reasoning to verify whether adjacent shots provide the relations required by individual criteria. For physical continuity, we retain pairs in which the same character, relevant object, or shared environmental landmark remains observable, enabling evaluation of spatial relations, object-state persistence, and relative scale. For cinematic grammar, shot-reverse-shot patterns support eyeline matching and 180-degree analysis, framing changes support shot progression, and hard cuts, dissolves, fades, or wipes support transition logic and rhythm.
\end{itemize}

Complete pipeline parameters, including detection thresholds, filtering criteria, and per-stage retention statistics, are detailed in Appendix B.

\noindent\textbf{Hierarchical Annotation.}
Each retained segment receives complementary segment- and shot-level annotations generated by a multimodal teacher. The \textbf{segment-level annotation} captures the narrative arc, character relationships, emotional trajectory, and cinematic structure, while the \textbf{shot-level annotations} describe character identity, physical state, facial affect, camera framing, interactions, and scene context. They further record cross-shot changes in spatial layout, object state, relative scale, gaze direction, screen direction, and transition type. We additionally retain native audio, source prompts, per-shot first frames, and other conditioning information for diverse generation settings.

\noindent\textbf{\ours{} Statistics.}
PersonaShot contains 1,000 human-centric multi-shot segments, averaging 5.3 shots per segment and totaling over 5,000 annotated shots. As shown  Fig.~\ref{fig:data_statistics}(bottom), the benchmark covers diverse narrative themes, sequence lengths, and framing scales, providing varied scenarios for evaluating character continuity across actions, emotional states, and interactions.

These distributions support different evaluation objectives: shorter sequences emphasize local cross-shot transitions, while longer sequences enable sequence-level trajectory analysis. Close and medium shots provide detailed facial evidence for affective evaluation, whereas wider views preserve the spatial and geometric cues required for physical continuity and cinematic grammar.

\begin{figure*}[ht]
    \centering
    \includegraphics[width=\linewidth]{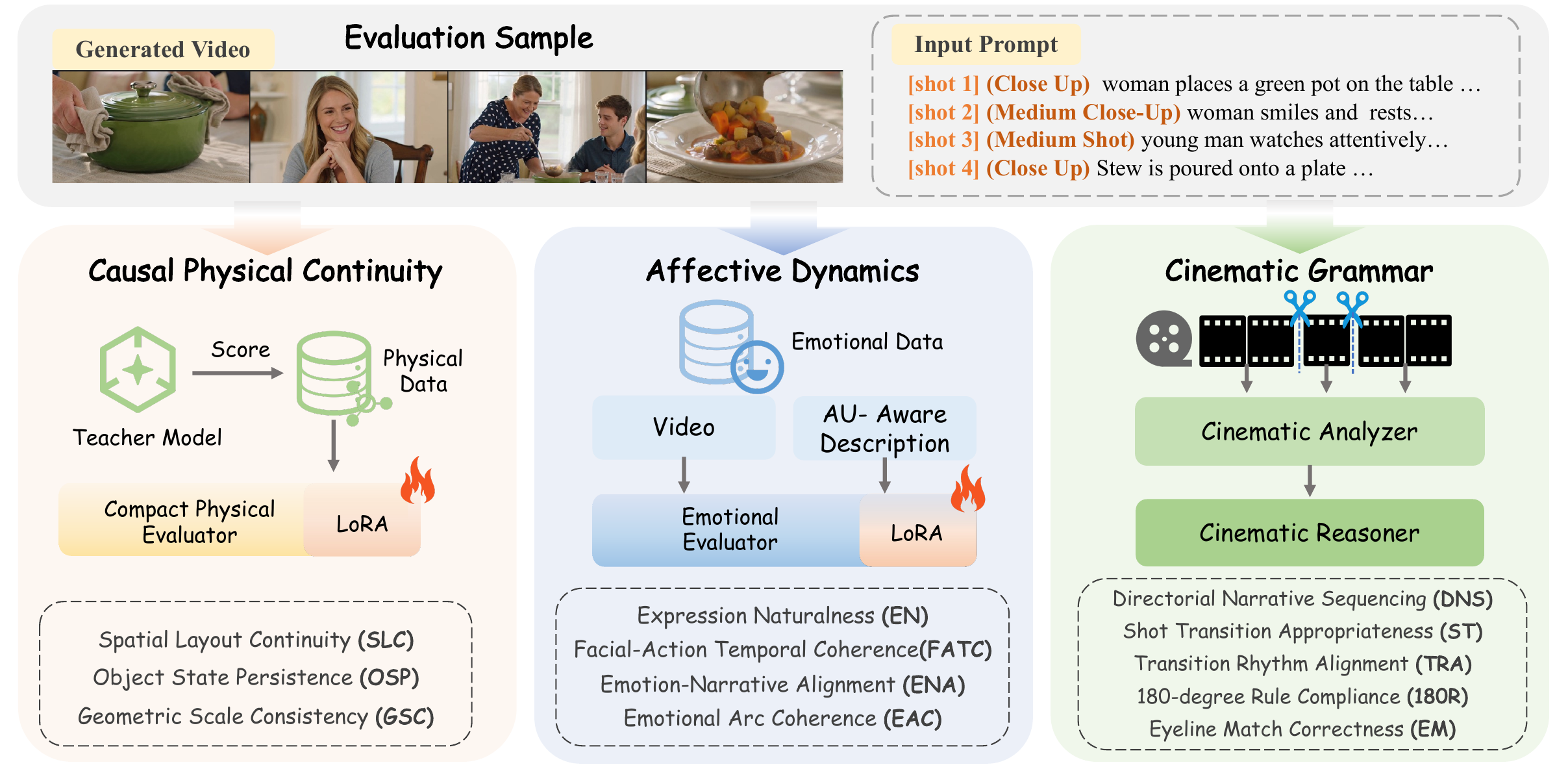}
    \caption{
\textbf{Overview of PersonaShot evaluation framework.}
Given generated multi-shot videos and prompts, PersonaShot extends conventional quality assessment with three specialist dimensions for narrative-level evaluation: causal physical continuity, affective dynamics, and cinematic grammar.
}
    \label{fig:method}
    \vspace{-3mm}
\end{figure*}

%% file: Secs/04_methodology.tex
\section{Fine-grained Evaluation Metrics}
\label{sec:method}

\subsection{Overview}

Existing video benchmarks mainly focus on perceptual quality, semantic alignment, or short-range consistency, providing limited support for failures emerging across editing cuts~\cite{bansal2025videophy,bansal2025videophy2,guo2026dreamid}. We therefore introduce \autoeval{}, a person-centric evaluation protocol for multi-shot video generation, organized along three temporal levels: within-shot states, cross-shot transitions, and sequence-level trajectories. The protocol contains 16 sub-metrics across four dimensions: visual quality and consistency, causal physical continuity, fine-grained affective dynamics, and cinematic grammar.

Given a generated sequence with segment- and shot-level annotations, we sample five representative shots along the timeline. Individual shots assess local character states, adjacent shot pairs evaluate cross-shot relations, and the full sequence measures narrative trajectories. Since different criteria require different evidence, \autoeval{} combines conventional perceptual models with three specialist evaluators that analyze physical states, affective dynamics, and cinematic structures using criterion-specific visual, facial, and temporal evidence.

\subsection{Visual Quality and Consistency}
\noindent\textbf{Dimension and Strategy.}
This foundational dimension assesses visual stability through four criteria: \textbf{1) Visual Fidelity (VF)}, \textbf{2) Text-Video Alignment (TVA)}, \textbf{3) Identity Consistency (ID)}, and \textbf{4) Scene/Style Consistency (SS)}. 
Following established protocols, we evaluate VF using DOVER and MUSIQ~\cite{wu2023dover,ke2021musiq}, TVA using VQAScore and GroundingDINO~\cite{lin2024vqascore,liu2024grounding}, and ID/SS via cross-shot embedding and perceptual similarities.

\subsection{Causal Physical Continuity}

\noindent\textbf{Dimension.}
Existing physical benchmarks mainly evaluate whether individual frames or short clips satisfy local physical plausibility~\cite{bansal2025videophy,bansal2025videophy2}. However, locally plausible shots may still form inconsistent physical processes across editing cuts. This dimension evaluates cross-shot physical evolution through three criteria: \textbf{1) Spatial Layout Continuity (SLC):} measuring whether character and object positions remain consistent across shots; \textbf{2) Object State Persistence (OSP):} evaluating whether object states evolve in a causally consistent manner; and \textbf{3) Geometric Scale Consistency (GSC):} measuring whether character-object proportions and scene geometry remain stable across viewpoints.

\noindent\textbf{Evaluation Strategy.}
To ensure genuine cross-shot reasoning rather than parsing ground-truth texts, we develop a specialist physical evaluator via teacher-guided distillation with strict input isolation. During training, a large multimodal teacher, Qwen3.5-397B-A17B~\cite{team2026qwen3}, receives visual evidence and structural annotations $\mathcal{A}$ to generate structured targets $y_{teacher}$. The compact Qwen3.5-4B evaluator $f_{\theta}$ (with LoRA adaptation~\cite{hu2022lora,bai2023qwen}) is optimized via:
\begin{equation}
\mathcal{L}_{distill} = \mathcal{L}_{CE}\big(f_{\theta}(\mathcal{V}_{gen}, \mathcal{P}_{text}), y_{teacher}\big),
\end{equation}
where annotations $\mathcal{A}$ are strictly isolated during inference, ensuring the evaluator relies solely on generated visual sequences $\mathcal{V}_{gen}$ and prompts $\mathcal{P}_{text}$ (more experimental details and training splits are deferred to Appendix B).

\subsection{Affective Dynamics}

\noindent\textbf{Dimension.}
Existing human-centric benchmarks commonly evaluate emotions using coarse categorical labels~\cite{cheng2024emotion}. However, such labels cannot capture subtle facial variations, temporal evolution, or narrative appropriateness of affective states. This dimension models emotional continuity through four criteria: \textbf{1) Expression Naturalness (EN):} evaluating facial realism and generation artifacts; \textbf{2) Facial-Action Temporal Coherence (FATC):} capturing facial-action stability and abrupt or static expressions; \textbf{3) Emotion-Narrative Alignment (ENA):} measuring whether emotional intensity and category match the current event; and \textbf{4) Emotional Arc Coherence (EAC):} evaluating whether affective states evolve coherently throughout the narrative.

\noindent\textbf{Evaluation Strategy.}
We adapt Qwen3.5-4B as an affective evaluator using emotion data from Emotion-LLaMA~\cite{cheng2024emotion}, trained via a similar distillation objective. To eliminate evaluation circularity, inference relies strictly on the generated video and extracted features rather than ground-truth labels. Specifically, we provide frame-level Action Unit (AU) signals extracted by OpenFace and formulate AU-aware evaluation instructions based on the Facial Action Coding System~\cite{baltruvsaitis2016openface}:
\begin{equation}
S_{aff} = f_{\theta}\big(\mathcal{V}_{gen}, \mathcal{P}_{text}, \mathcal{F}_{AU}\big),
\end{equation}
where $\mathcal{F}_{AU}$ denotes the frame-level AU temporal dynamics, enabling the evaluator to jointly reason over facial observations and narrative context without annotation leakage (see Appendix B for further implementation specifics).

\subsection{Cinematic Grammar}
\noindent\textbf{Dimension.}
Existing benchmarks~\cite{wei2026msavbench,shi2026msvbench} may evaluate camera motion or shot-level properties, but rarely measure whether generated sequences follow established cinematic conventions~\cite{wang2026cinetechbench}. Cinematic grammar therefore evaluates five complementary aspects: \textbf{1) Directorial Narrative Sequencing (DNS):} assessing whether a global prompt is translated into a coherent shot-level narrative structure; \textbf{2) Shot Transition Appropriateness (ST):} determining whether transition types match the narrative context; \textbf{3) Transition Rhythm Alignment (TRA):} measuring whether shot duration and cutting density correspond to narrative pacing; \textbf{4) 180-degree Rule Compliance (180R):} examining whether screen direction and action axes remain consistent across cuts; and \textbf{5) Eyeline Match Correctness (EM):} evaluating whether gaze relations are spatially coherent between shots.

\noindent\textbf{Evaluation Strategy.}
We model cinematic grammar through a two-stage process consisting of cinematic analysis and narrative reasoning. 
First, we employ cinematographic video captioning models~\cite{mao2026cinecap} to extract professional film-language descriptions, including shot scale, camera perspective, composition, and editing cues, from generated sequences. 
Shot boundaries and transition structures are further identified using TransNetV2~\cite{soucek2024transnet}. 
Second, a cinematic reasoner evaluates whether the extracted cinematic structures satisfy narrative and editing principles.
DNS is evaluated by comparing generated shot structures with the intended global narrative, while ST and TRA are assessed by jointly considering transition patterns, shot duration, and narrative context. 
Spatial cinematic rules, including 180R and EM, are evaluated through character orientation, screen direction, and gaze consistency across adjacent shots.

\subsection{Score Aggregation}
The heterogeneous metric outputs are first converted into higher-is-better scores and normalized to $[0,1]$ using metric-specific calibration functions. The overall PersonaShot score is computed as:

\begin{equation}
S_{\mathrm{PS}}
=
\alpha_{\mathrm{vis}}S^{\mathrm{vis}}
+
\alpha_{\mathrm{phy}}S^{\mathrm{phy}}
+
\alpha_{\mathrm{aff}}S^{\mathrm{aff}}
+
\alpha_{\mathrm{cin}}S^{\mathrm{cin}},
\end{equation}
where $S^{\mathrm{vis}}$, $S^{\mathrm{phy}}$, $S^{\mathrm{aff}}$, and $S^{\mathrm{cin}}$ denote visual quality and consistency, causal physical continuity, fine-grained affective dynamics, and cinematic grammar scores, respectively. The coefficients $\alpha_{\mathrm{vis}}$, $\alpha_{\mathrm{phy}}$, $\alpha_{\mathrm{aff}}$, and $\alpha_{\mathrm{cin}}$ are fixed benchmark hyperparameters that balance foundational visual quality with the three narrative continuity dimensions.

%% file: Secs/05_experiments.tex
\section{Experiments}
\label{sec:experiments}

\input{Tabs/main_exp}
\subsection{Experimental Setup}
We evaluate representative multi-shot video generation systems on the same 1,000 benchmark samples. The compared methods cover both \textit{shot-level prompt conditioning} and \textit{global-prompt conditioning}, including systems with explicit shot planning and those generating shots independently.

Our benchmark comprises 16 sub-metrics in total across four dimensions. For a fair comparison, Table~\ref{tab:main_results} reports the 15 common metrics shared across all settings, while sequence-level Directorial Narrative Sequencing (DNS) is evaluated specifically for global-prompt narratives and discussed in Finding 5. All metric scores are normalized to $[0,1]$ before aggregation. Following expert preference voting and benchmark design considerations, we assign equal weights to the four dimensions, \textit{i.e.},
$\alpha_{\mathrm{vis}}=\alpha_{\mathrm{phy}}=\alpha_{\mathrm{aff}}=\alpha_{\mathrm{cin}}=0.25$.
More evaluation details are shown in Appendix C.

\subsection{Main Results and Insights}

Table~\ref{tab:main_results} summarizes the fine-grained evaluation results of representative multi-shot video generation systems. Beyond the overall ranking, \ours{} reveals several important characteristics of current models.

\noindent\textbf{Finding 1: Perceptual quality does not guarantee narrative continuity.}
Modern video generators have achieved strong visual fidelity, but substantial gaps remain in maintaining character-centric coherence across shots. For example, ShotStream obtains high visual fidelity yet a lower aggregate score due to limited performance in affective and cinematic dimensions. Similarly, several models with competitive visual quality exhibit significant degradation in physical state persistence and emotional evolution, indicating that conventional evaluation can overestimate multi-shot generation capabilities.

\noindent\textbf{Finding 2: Explicit shot planning improves structural coherence, but global narrative reasoning remains essential.}
Models with shot-level conditioning generally achieve stronger cross-shot consistency via explicit guidance. Seedance achieves the leading overall performance under both settings. However, global-prompt systems such as LTX-2.3 remain competitive in cinematic grammar and identity, suggesting that effective narrative planning matters more than prompt granularity alone.

\noindent\textbf{Finding 3: Physical and affective state propagation remain major challenges.}
Although current models generate visually plausible individual shots, they struggle to preserve latent character and scene states across editing boundaries. Physical continuity scores reveal frequent failures in object-state persistence and spatial consistency, while affective evaluation exposes unstable facial dynamics and weak emotional trajectories throughout multi-shot sequences.

\noindent\textbf{Finding 4: Cinematic grammar is largely under-modeled in existing video generators.}
\ours{} further reveals that visually coherent shots do not necessarily form cinematographically valid sequences. Many systems struggle with transition appropriateness, rhythm alignment, 180-degree rule compliance, and eyeline matching, indicating that current models primarily optimize visual synthesis rather than film-language reasoning.

\noindent\textbf{Finding 5: Global narrative sequencing reveals a stark divide in long-form cinematic control.}
Evaluating global-prompt settings allows us to isolate Directorial Narrative Sequencing (DNS), measuring whether a global prompt translates into a coherent structure. Here, Seedance achieves $0.865$ in DNS, whereas open-source LTX-2.3 and VGoT achieve $0.624$ and $0.535$. This performance gap highlights that mastering long-form narrative planning remains a key bottleneck for open-source models.

\begin{figure*}
    \centering
    \includegraphics[width=\linewidth]{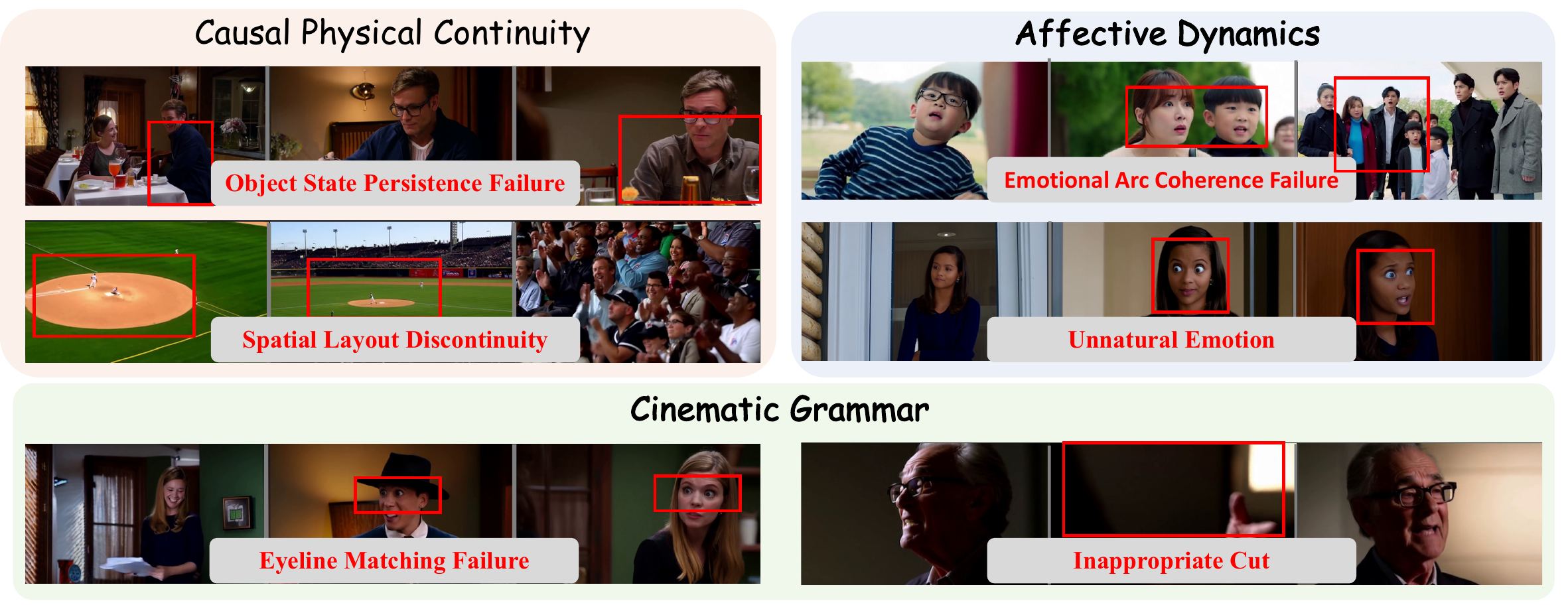}
    \caption{\textbf{Qualitative failure cases revealed by \ours{}.}
The examples illustrate representative failures in causal physical continuity, affective dynamics, and cinematic grammar, including object state persistence failures, spatial layout discontinuities, emotion evolution inconsistencies, facial expression issues, eyeline matching failures, and inappropriate shot transitions.}
    \label{fig:failure}
\end{figure*}

\subsection{Failure Case Analysis}

Beyond quantitative comparisons, \ours{} provides interpretable diagnoses of multi-shot generation failures. 
As illustrated in Fig.~\ref{fig:failure}, existing models frequently fail to preserve object states and spatial relations across shots, maintain coherent emotional evolution, and follow basic cinematic conventions such as eyeline matching and transition consistency. 
These failures are often visually plausible at the individual-shot level but become apparent when evaluated across temporal and relational dimensions. 
The results highlight that reliable multi-shot generation requires persistent reasoning over physical states, affective trajectories, and cinematic structures beyond conventional visual quality.

\subsection{Human Study and Evaluator Validation}

To assess whether the proposed evaluators reflect human judgments, we conduct an expert study on 25 diverse generated multi-shot sequences. Ten domain experts participate under a sparse assignment protocol, with each sequence independently rated by four annotators using criterion-specific five-point rubrics. The ratings are averaged into Mean Opinion Scores (MOS), and the resulting Krippendorff's $\alpha=0.76$ indicates substantial inter-rater agreement. We measure alignment using Spearman correlation ($\rho$) and pairwise ranking agreement. Complete annotation rubrics, valid sample counts, confidence intervals, reliability statistics, and results for all 12 specialist criteria are provided in the Appendix E.

As shown in Table~\ref{tab:human_alignment}, the proposed evaluators consistently align with human judgments across all three specialist dimensions. The aggregated physical, affective, and cinematic scores achieve Spearman correlations of 0.75, 0.69, and 0.73, respectively, with pairwise agreement between 70.3\% and 75.2\% (all $p<0.01$). Object State Persistence shows strong alignment ($\rho=0.74$), reflecting its explicit cross-shot evidence. Emotion-Narrative Alignment also correlates well with human ratings ($\rho=0.71$), indicating sensitivity to the narrative appropriateness of facial affect. Facial-Action Temporal Coherence is comparatively more challenging because subtle facial changes are inherently ambiguous. The cinematic results further show that transition and eyeline relations can be evaluated consistently, supporting the complementary diagnostic value of the three dimensions.

\begin{table}[t]
\centering

\footnotesize
\setlength{\tabcolsep}{4.2pt}
\renewcommand{\arraystretch}{1.06}

\begin{tabular}{@{}llcc@{}}
\toprule
\textbf{Dimension} & \textbf{Criterion} & $\boldsymbol{\rho}\uparrow$ & \textbf{Pair.}$\uparrow$ \\
\midrule

\multirow{3}{*}{Physical} & Spatial Layout (SL) & 0.68 & 71.4\% \\
& Object State (OSP) & 0.74 & 76.0\% \\
& \textbf{Aggregate} & \textbf{0.75} & \textbf{75.2\%} \\
\addlinespace[2pt]

\multirow{3}{*}{Affective} & Emotion Alignment (ENA) & 0.71 & 72.5\% \\
& Facial Dynamics (FATC) & 0.68 & 64.8\% \\
& \textbf{Aggregate} & \textbf{0.69} & \textbf{70.3\%} \\
\addlinespace[2pt]

\multirow{3}{*}{Cinematic} & Shot Transition (ST) & 0.72 & 73.1\% \\
& Eyeline Match (EM) & 0.65 & 67.2\% \\
& \textbf{Aggregate} & \textbf{0.73} & \textbf{72.8\%} \\

\bottomrule
\end{tabular}
\caption{
\textbf{Human alignment of \ours{} evaluators.} We report Spearman correlation ($\rho$) with human Mean Opinion Scores ($N=25$) and pairwise agreement (Pair.) for representative criteria. Full results are in the Appendix E.
}
\vspace{-4mm}
\label{tab:human_alignment}
\end{table}

%% file: Tabs/main_exp.tex
\begin{table*}[t]
\centering

\scriptsize
\setlength{\tabcolsep}{2.25pt}
\renewcommand{\arraystretch}{0.98}

\resizebox{\textwidth}{!}{
\begin{tabular}{
@{}l
*{4}{c}@{\hspace{3pt}}
*{3}{c}@{\hspace{3pt}}
*{4}{c}@{\hspace{3pt}}
*{4}{c}@{\hspace{3pt}}
c@{}
}
\toprule

\multirow{2}{*}{\textbf{Method}}
& \multicolumn{4}{c}{\textbf{Visual Quality and Consistency}}
& \multicolumn{3}{c}{\textbf{Causal Physical Continuity}}
& \multicolumn{4}{c}{\textbf{Affective Dynamics}}
& \multicolumn{4}{c}{\textbf{Cinematic Grammar}}
& \multirow{2}{*}{\textbf{Aggregate}}
\\

\cmidrule(lr){2-5}
\cmidrule(lr){6-8}
\cmidrule(lr){9-12}
\cmidrule(lr){13-16}

& \textbf{VF}
& \textbf{TVA}
& \textbf{ID}
& \textbf{SS}
& \textbf{SLC}
& \textbf{OSP}
& \textbf{GSC}
& \textbf{EN}
& \textbf{FATC}
& \textbf{ENA}
& \textbf{EAC}
& \textbf{ST}
& \textbf{TRA}
& \textbf{180R}
& \textbf{EM}
&
\\

\midrule

\multicolumn{17}{@{}l}{\textit{\textbf{Shot-Level Prompt Conditioning}}}
\\[1pt]

Seedance
& \underline{0.981}
& \underline{0.752}
& \textbf{0.673}
& \textbf{0.762}
& \textbf{0.778}
& \textbf{0.928}
& \textbf{0.947}
& \textbf{0.907}
& \textbf{0.562}
& \textbf{0.778}
& \textbf{0.617}
& \textbf{0.837}
& \textbf{0.658}
& \textbf{0.818}
& \textbf{0.813}
& \textbf{0.793}
\\

HoloCine
& 0.853
& \textbf{0.797}
& \underline{0.582}
& 0.718
& 0.647
& \underline{0.862}
& \underline{0.893}
& 0.792
& 0.384
& \underline{0.677}
& 0.343
& \underline{0.723}
& 0.517
& \underline{0.712}
& 0.687
& \underline{0.687}
\\

STAGE
& 0.827
& 0.703
& 0.507
& 0.752
& 0.623
& 0.818
& 0.852
& 0.718
& 0.453
& 0.628
& \underline{0.443}
& 0.677
& 0.503
& 0.667
& 0.642
& 0.661
\\

LTX-2.3
& 0.892
& 0.718
& 0.572
& 0.733
& 0.677
& 0.832
& 0.857
& 0.758
& 0.427
& 0.583
& 0.403
& 0.683
& 0.512
& 0.683
& 0.648
& 0.673
\\

MultiShotMaster
& 0.849
& 0.697
& 0.557
& 0.653
& 0.604
& 0.852
& 0.884
& 0.781
& \underline{0.458}
& 0.596
& 0.442
& 0.622
& 0.482
& 0.623
& 0.603
& 0.655
\\

LongLive
& 0.972
& 0.458
& 0.532
& \underline{0.753}
& \underline{0.697}
& 0.697
& 0.803
& 0.687
& 0.383
& 0.418
& 0.348
& 0.684
& 0.468
& 0.697
& 0.687
& 0.626
\\

ShotStream
& \textbf{0.994}
& 0.377
& 0.548
& 0.673
& 0.668
& 0.573
& 0.742
& 0.677
& 0.387
& 0.383
& 0.367
& 0.688
& 0.497
& 0.692
& \underline{0.693}
& 0.601
\\

CineTrans
& 0.795
& 0.557
& 0.538
& 0.682
& 0.572
& 0.697
& 0.708
& 0.613
& 0.378
& 0.503
& 0.358
& 0.594
& \underline{0.537}
& 0.605
& 0.587
& 0.586
\\

EchoShot
& 0.937
& 0.507
& 0.463
& 0.488
& 0.482
& 0.758
& 0.871
& \underline{0.802}
& 0.453
& 0.448
& 0.432
& 0.503
& 0.478
& 0.488
& 0.482
& 0.581
\\

\midrule

\multicolumn{17}{@{}l}{\textit{\textbf{Global-Prompt Conditioning}}}
\\[1pt]

Seedance
& \underline{0.973}
& \textbf{0.712}
& 0.642
& \underline{0.733}
& \textbf{0.738}
& \textbf{0.908}
& \textbf{0.937}
& \textbf{0.882}
& \textbf{0.532}
& \textbf{0.743}
& \textbf{0.588}
& \textbf{0.808}
& \textbf{0.627}
& \textbf{0.788}
& \textbf{0.783}
& \textbf{0.766}
\\

LTX-2.3
& 0.818
& \underline{0.523}
& \textbf{0.697}
& \textbf{0.823}
& \underline{0.697}
& \underline{0.703}
& 0.758
& 0.683
& \underline{0.403}
& 0.453
& 0.348
& \underline{0.697}
& 0.472
& \underline{0.703}
& 0.677
& 0.636
\\

VGoT
& \textbf{0.988}
& 0.418
& \underline{0.683}
& 0.648
& 0.627
& 0.608
& \underline{0.881}
& \underline{0.868}
& 0.388
& \underline{0.477}
& \underline{0.357}
& 0.643
& \underline{0.557}
& 0.682
& \underline{0.683}
& \underline{0.638}
\\

\bottomrule
\end{tabular}
}
\caption{
\textbf{Fine-grained evaluation results on PersonaShot under
shot-level and global-prompt conditioning.}
Higher is better. The aggregate score equally averages the four
dimension scores, with metrics averaged within each dimension.
Within each protocol, the best and second-best results are highlighted
in \textbf{bold} and \underline{underlined}, respectively.
}
\label{tab:main_results}
\end{table*}

%% file: Secs/06_conclusion.tex
\section{Conclusion}
\label{sec:conclusion}

We introduced \ours{}, the first person-centric benchmark for evaluating narrative continuity in multi-shot video generation. Beyond conventional perceptual quality assessment, \ours{} systematically measures three critical dimensions of character-centric storytelling: causal physical continuity, fine-grained affective dynamics, and cinematic grammar. Through 16 complementary metrics and specialist evaluators aligned with human judgments, our benchmark reveals that current video generators remain limited in preserving persistent character states, modeling emotional evolution, and following cinematic conventions across shots.

Our findings highlight a key direction for future multi-shot generation: moving beyond isolated visually plausible shots toward coherent narrative systems that explicitly model physical states, affective trajectories, and cinematic structures. Furthermore, by providing lightweight yet human-aligned specialist evaluators, our framework offers a reliable and interpretable recipe for iterative model diagnosis. We hope \ours{} can serve as both a rigorous testbed and a catalyst for developing next-generation video foundation models with authentic storytelling capabilities.

%% file: Suppl_secs/detailed_related_work.tex
\section{Detailed Related Work}
\label{app:related_work}

\subsection{Video Generation Benchmarks}

Early video generation benchmarks primarily focused on general-purpose, single-shot scenarios, establishing comprehensive hierarchical dimensions for isolated clips~\cite{huang2024vbench,liu2024evalcrafter}. Concurrently, specialized person-centric benchmarks, such as HVEval~\cite{wu2025hveval} and HumanVBench~\cite{zhou2024humanvbench}, began to address human-specific criteria, though they evaluate either single-shot generation quality or video understanding capabilities, without measuring cross-shot character coherence. As generation capabilities advanced, recent frameworks have shifted toward multi-shot evaluations. Benchmarks including MSVBench~\cite{shi2026msvbench}, UniVBench~\cite{wei2026univbench}, and MSAVBench~\cite{wei2026msavbench} introduced metrics for narrative consistency, agent-based scoring, and audio-visual synchronization to assess longer, multi-cut sequences. MuSS~\cite{zhang2026muss} introduces cinematic narrative evaluation but focuses on subject-to-video generation without person-centric cross-shot analysis. ViStoryBench~\cite{zhuang2026vistorybench} evaluates story visualization quality but does not assess character state continuity across cuts.

However, these multi-shot frameworks lack fine-grained analysis in dimensions crucial for cross-shot coherence. Specifically, they overlook physical rationality across shot boundaries, reduce emotional evaluation to coarse categorical labels (e.g., basic emotion categories as in Emotion-LLaMA~\cite{cheng2024emotion}) rather than tracking micro-expression dynamics, and entirely ignore rule-based cinematic grammar. To bridge these gaps, \ours{} provides a unified, person-centric benchmark that integrates cross-shot physical consistency, Action Unit (AU)-level micro-expression analysis, and cinematic grammar quantification.

\subsection{Multi-Shot Video Generation}
Multi-shot video generation has flourished recently, driven by both open-source efforts and commercial systems. Models differ primarily in how much structure users must provide. In \textbf{manual per-shot} generation, users write a dedicated prompt for each shot: EchoShot~\cite{wang2026echoshot} accepts structured shot-level captions, while ShotStream~\cite{luo2026shotstream} and LongLive~\cite{yang2025longlive} support streaming interactive input with KV-cache continuity. \textbf{Global plus per-shot} methods, including MultiShotMaster~\cite{wang2026multishotmaster}, HoloCine~\cite{meng2026holocine}, and CineTrans~\cite{wu2025cinetrans}, let users supply an overarching scene description alongside per-shot captions. \textbf{Auto-decomposition} approaches, STAGE~\cite{zhang2026stage} and VGoT~\cite{zheng2024vgot}, accept only a high-level story theme or a single sentence and automatically generate a structured storyboard via a director agent or GPT-4o-based planning. \textbf{Flexible-conditioning} models such as LTX-2.3~\cite{hacohen2026ltx} and Seedance~\cite{seedance2026seedance} adapt to both shot-level and global-prompt settings, handling shot decomposition either internally or guided by explicit per-shot prompts. We benchmark these representative methods under a unified protocol to assess current multi-shot capabilities and identify key directions for future improvements.

\subsection{Evaluation with Vision-Language Models}

Scaling evaluation beyond human annotation relies increasingly on Vision-Language Models (VLMs). However, zero-shot VLMs exhibit surface-level bias, conflating visual quality with structural correctness and missing specific cross-shot violations~\cite{bansal2025videophy}. Task-specific fine-tuning mitigates this, as seen in VideoPhy-2~\cite{bansal2025videophy2} for physical commonsense, Emotion-LLaMA~\cite{cheng2024emotion} for emotion recognition, and CineCap~\cite{mao2026cinecap} for cinematic captioning. \ours{} advances this paradigm via teacher-guided distillation: we train multiple lightweight specialized evaluators from a Qwen3.5-397B-A17B teacher~\cite{team2026qwen3}, each grounded in the criterion-specific evidence required by its metric---visual and temporal cues for the physical continuity evaluator, AU-signaled facial dynamics for the affective evaluator, and structured cinematic descriptions for the two-stage cinematic reasoner. Validated by an expert study with 10 domain specialists, these distilled evaluators achieve substantial alignment with human judgments (Spearman $\rho$ up to 0.75, all $p < 0.01$), delivering scalable, expert-level benchmarking.

%% file: Suppl_secs/implementation_detail.tex
\begin{table*}[t]
\centering
\small
\caption{\textbf{LoRA Fine-Tuning Configurations for Specialist Evaluators.} Training setup and hyperparameters for adapting Qwen3-4B and Qwen3-8B on the 4,041-sample MERR training set.}
\label{tab:lora_hyperp}
\begin{tabular}{l c c}
\toprule
\textbf{Hyperparameter} & \textbf{Qwen3-4B Evaluator} & \textbf{Qwen3-8B Evaluator} \\
\midrule
Base Model Backbone & \texttt{Qwen/Qwen3-4B} & \texttt{Qwen/Qwen3-8B} \\
Training / Validation Samples & \multicolumn{2}{c}{4,041 / 446 (MERR dataset)} \\
Target Emotion Categories & \multicolumn{2}{c}{9 classes} \\
LoRA Rank ($r$) & 16 & 32 \\
LoRA Alpha ($\alpha$) & 32 & 64 \\
Trainable Parameters & 33M (0.81\%) & 66M ($\sim$0.80\%) \\
Training Epochs & 3 & 10 \\
Learning Rate & $2 \times 10^{-4}$ & $1 \times 10^{-4}$ \\
Batch Size (per device) & 4 & 4 \\
Training Duration & $\sim$33 min & $\sim$42 min \\
\bottomrule
\end{tabular}
\end{table*}

\section{Data Processing \& Evaluator Training Details}
\label{app:data_and_training}

In this section, we provide the complete technical implementation details for \ours{}, including the automated data curation pipeline and its retention statistics, as well as the training datasets, hyperparameters, and inference protocols for our specialist evaluators.

\subsection{Automatic Data Processing \& Pipeline Details}
\label{app:data_processing}

To construct \ours{}, we source raw video sequences from two large-scale corpora: CineDance-1M~\cite{chen2026cinedance} and OpenHumanVid~\cite{li2025openhumanvid}. A three-stage automated filtering pipeline is applied to select segments with sufficient visual, facial, and relational evidence for person-centric multi-shot evaluation.

\paragraph{Stage 1: Sequence Segmentation and Character Continuity.}
We use TransNet V2~\cite{soucek2024transnet} to detect shot transitions and slice raw videos into candidate multi-shot segments ranging from 2 to 15 shots (with an average length of 5.3 shots). To enforce narrative coherence, we require recurring human anchor identities: segments are retained only if at least one primary character appears across multiple shots within a continuous local event.

\begin{table*}[t]
\centering
\small
\setlength{\tabcolsep}{4pt}
\renewcommand{\arraystretch}{1.15}
\begin{tabularx}{\textwidth}{
    >{\raggedright\arraybackslash}p{0.18\textwidth}
    >{\raggedright\arraybackslash}p{0.20\textwidth}
    >{\raggedright\arraybackslash}p{0.27\textwidth}
    >{\raggedright\arraybackslash}X
}
\toprule
\textbf{Dimension} & \textbf{Metric} & \textbf{Definition} & \textbf{Evaluation Strategy} \\
\midrule
\multicolumn{4}{l}{\textbf{Visual Quality and Consistency}} \\
\midrule
Visual Quality and Consistency & \textbf{Visual Fidelity (VF)} & Measures perceptual quality, artifacts, distortions, and composition. & DOVER aesthetic/technical assessment and MUSIQ frame-level quality evaluation. \\
 & \textbf{Text-Video Alignment (TVA)} & Measures whether generated content follows textual instructions, including entities, actions, and scenes. & VQAScore semantic matching combined with GroundingDINO-based entity grounding. \\
 & \textbf{Identity Consistency (ID)} & Measures whether recurring characters preserve visual identity across shot transitions. & Character localization and tracking followed by cross-shot identity embedding similarity. \\
 & \textbf{Scene \& Style Consistency (SS)} & Measures background, lighting, color, and visual style stability across shots. & Background similarity, style embedding comparison, and multimodal visual consistency judgments. \\
 & \textbf{Reference Fidelity (RF)} & Measures preservation of identity from reference images or audio conditions. & DINOv2 and ArcFace similarities for visual identity preservation. \\
\midrule
\multicolumn{4}{l}{\textbf{Causal Physical Continuity}} \\
\midrule
Causal Physical Continuity & \textbf{Spatial Layout Continuity (SLC)} & Measures whether character and object positions remain consistent across cuts. & Character/object localization and tracking evaluate normalized spatial displacement between adjacent shots. \\
 & \textbf{Object State Persistence (OSP)} & Evaluates whether object states evolve causally instead of being reset across shots. & A multimodal evaluator extracts object states and verifies physically plausible state transitions. \\
 & \textbf{Geometric Scale Consistency (GSC)} & Measures whether character-object proportions and scene geometry remain stable across viewpoints. & Depth estimation and geometric reasoning evaluate relative scale consistency. \\
\bottomrule
\end{tabularx}
\caption{\textbf{Detailed evaluation metrics of PersonaShot (Part I).} The table summarizes the foundational visual metrics and causal physical continuity metrics.}
\label{tab:metric_details_part1}
\end{table*}

\paragraph{Stage 2: Face Visibility and Quality Filtering.}
For fine-grained affective evaluation, we sample candidate frames uniformly across each shot and detect faces using RetinaFace~\cite{deng2020retinaface}. We apply a strict confidence threshold of $\ge 0.85$ and require a minimum face bounding box area of $\ge 5\%$ of the total frame. A segment is retained for affective evaluation only when valid face crops are present in at least $60\%$ of its constituent shots and across no fewer than two distinct shots. To filter out severe motion blur or generative artifacts, we apply CLIP-IQA~\cite{wang2023clipiqa} and discard face crops scoring in the bottom 15th percentile of the data distribution.

\paragraph{Stage 3: Cross-Shot Relation Selection \& Benchmark Statistics.}
We employ Qwen3.5-397B-A17B~\cite{team2026qwen3} to inspect adjacent shot pairs and verify the presence of required relational cues:
\begin{itemize}
    \item \textbf{Physical Continuity Evidence:} Adjacent shots must retain observable commonalities, including identical character identities, interacted props, or consistent background landmarks, to support evaluations of Spatial Layout Continuity (\textbf{SLC}), Object State Persistence (\textbf{OSP}), and Geometric Scale Consistency (\textbf{GSC}).
    \item \textbf{Cinematic Grammar Evidence:} Shot transitions are categorized (hard cuts, dissolves, fades, and wipes) via TransNet V2, and shot-reverse-shot structures are tagged to support 180-degree Rule Compliance (\textbf{180R}) and Eyeline Match Correctness (\textbf{EM}).
\end{itemize}
After completing the three filtering stages, the final benchmark comprises exactly 1,000 high-quality human-centric multi-shot video segments, totaling over 5,000 annotated shots across diverse narrative themes.

\paragraph{Three-Layer Evaluation Workflow.}
During automated benchmarking, inference is structured across three temporal layers to minimize computational overhead:
\begin{enumerate}
    \item \textbf{Shot-Layer:} For 5 representative shots per video (sampled at the 0\%, 25\%, 50\%, 75\%, and 100\% timeline intervals), we perform a unified 7-in-1 visual scoring call and an Emotion-Narrative Alignment call on the middle keyframe (10 calls per video).
    \item \textbf{Cross-Shot Layer:} We perform 2 pairwise image-comparison calls on consecutive keyframe pairs (evaluating spatial, eyeline, and 180-degree rules) and 2 text-based calls on the extracted emotion label sequence (evaluating temporal flow and emotional arc).
    \item \textbf{Sequence-Layer:} Transition Rhythm Alignment (\textbf{TRA}) is computed programmatically by combining duration variation statistics with the VLM-extracted action and emotion intensity scores.
\end{enumerate}

\subsection{Evaluator Distillation, Fine-Tuning, and Inference Setup}
\label{app:evaluator_training}

To deploy lightweight, human-aligned specialist evaluators locally, we adapt Qwen3.5-4B and Qwen3-8B models via Low-Rank Adaptation (LoRA)~\cite{hu2022lora}.

\paragraph{Training Data Scale \& Emotion Fine-Tuning Setup.}
For our fine-grained affective evaluator and Emotion-Prompt Alignment (\textbf{EPA}) scoring model, we fine-tune the Qwen3 backbone on the Multimodal Emotion Recognition and Reasoning (MERR) dataset. The training corpus consists of exactly \textbf{4,041 training samples} and \textbf{446 validation samples} covering 9 distinct emotional categories (\textit{happy, sad, neutral, angry, worried, surprise, fear, doubt, and contempt}). 

During training, the model learns to map multimodal facial Action Unit (AU) descriptions and narrative contexts to precise emotion labels. Table~\ref{tab:lora_hyperp} details the exact LoRA hyperparameters and training configurations used for both the 4B and 8B model variants.

\paragraph{Teacher-Guided Distillation \& Input Isolation Mechanism.}
For the physical continuity and cinematic grammar evaluators, we generate structured supervision targets $y_{\text{teacher}}$ using a Qwen3.5-397B-A17B teacher model conditioned on full segment annotations $\mathcal{A}$. To prevent the student evaluators from developing a shortcut dependency on ground-truth text annotations during inference, we enforce an input isolation mechanism:
\begin{equation}
    \mathcal{S}_{\text{phy}} = f_{\theta}(\mathcal{V}_{\text{gen}}, \mathcal{P}_{\text{text}}), \quad \mathcal{S}_{\text{aff}} = f_{\theta}(\mathcal{V}_{\text{gen}}, \mathcal{P}_{\text{text}}, \mathcal{F}_{\text{AU}}),
\end{equation}
where $\mathcal{A}$ is strictly withheld during student evaluation. Here, $\mathcal{F}_{\text{AU}}$ represents frame-level Action Unit features extracted using OpenFace~\cite{baltruvsaitis2016openface} formatted under the Facial Action Coding System (FACS), ensuring that the evaluator scores facial dynamics purely from generated visual evidence.

%% file: Suppl_secs/detailed_metrics.tex
\begin{table*}[t]
\centering
\small
\setlength{\tabcolsep}{4pt}
\renewcommand{\arraystretch}{1.15}
\begin{tabularx}{\textwidth}{
    >{\raggedright\arraybackslash}p{0.18\textwidth}
    >{\raggedright\arraybackslash}p{0.20\textwidth}
    >{\raggedright\arraybackslash}p{0.27\textwidth}
    >{\raggedright\arraybackslash}X
}
\toprule
\textbf{Dimension} & \textbf{Metric} & \textbf{Definition} & \textbf{Evaluation Strategy} \\
\midrule
\multicolumn{4}{l}{\textbf{Affective Dynamics}} \\
\midrule
Affective Dynamics & \textbf{Expression Naturalness (EN)} & Measures facial realism and generation artifacts in expressions. & Multimodal expression assessment combined with facial Action Unit (AU) analysis. \\
 & \textbf{Facial-Action Temporal Coherence (FATC)} & Measures temporal stability of facial actions and detects jitter or unnatural static expressions. & Frame-level AU trajectories are analyzed through temporal variation statistics. \\
 & \textbf{Emotion-Narrative Alignment (ENA)} & Measures whether emotional category and intensity match the narrative context. & Narrative affect extracted from annotations is compared with AU-aware facial affect estimation. \\
 & \textbf{Emotional Arc Coherence (EAC)} & Measures whether emotional states evolve coherently throughout the sequence. & Affective trajectories are constructed across shots and compared with expected narrative evolution. \\
\midrule
\multicolumn{4}{l}{\textbf{Cinematic Grammar}} \\
\midrule
Cinematic Grammar & \textbf{Directorial Narrative Sequencing (DNS)} & Measures whether a global prompt is translated into a coherent shot-level narrative structure. & Global prompts and generated sequences are evaluated through cinematic reasoning models. \\
 & \textbf{Shot Transition Appropriateness (ST)} & Measures whether transition types match narrative context. & Shot boundaries are detected and transition choices are evaluated with visual and narrative evidence. \\
 & \textbf{Transition Rhythm Alignment (TRA)} & Measures whether shot duration and cutting density correspond to narrative pacing. & Shot duration statistics and cutting patterns are compared with narrative tension. \\
 & \textbf{180-degree Rule Compliance (180R)} & Measures whether screen direction and action axes remain consistent across cuts. & Character orientation and cross-shot axis consistency are analyzed. \\
 & \textbf{Eyeline Match Correctness (EM)} & Measures whether gaze directions are spatially coherent in shot-reverse-shot structures. & Facial gaze estimation evaluates complementary eye-line relations across shots. \\
\bottomrule
\end{tabularx}
\caption{\textbf{Detailed evaluation metrics of PersonaShot (Part II).} The table summarizes affective dynamics and cinematic grammar metrics.}
\label{tab:metric_details_part2}
\end{table*}

\section{Detailed Metrics}
To provide a comprehensive operational description of PersonaShot, we summarize the definition and evaluation strategy for all 16 fine-grained metrics in Tables~\ref{tab:metric_details_part1} and \ref{tab:metric_details_part2}. The evaluation suite is structured across four complementary dimensions: foundational visual quality and consistency, causal physical continuity, affective dynamics, and cinematic grammar.

\paragraph{Evidence-Driven Dynamic Activation.}
To avoid inappropriate penalization, each metric is dynamically activated only when sufficient visual, facial, or relational evidence is verified by our automated preprocessing pipeline (\S B.1). For example, eyeline matching (\textbf{EM}) is evaluated strictly on identified shot-reverse-shot structures, whereas Action Unit-aware affective metrics require clear facial observations (\textbf{FATC}, \textbf{ENA}). When a specific relational cue is absent in a video segment, the corresponding metric is masked out, and the dimension score is computed as the normalized average over all valid evaluation instances.

\paragraph{Score Calibration and Aggregation.}
Since the 16 sub-metrics originate from heterogeneous evaluation engines---ranging from DOVER quality scores and similarity cosines to 1-to-5 Likert ratings from specialist evaluators---all raw outputs are first calibrated to a standardized $[0, 1]$ interval via metric-specific linear scaling, where higher values consistently indicate better narrative continuity. Following the standard evaluation protocol established in the main paper, the aggregate PersonaShot score ($S_{\text{PS}}$) is computed as the unweighted arithmetic mean of the four dimension scores ($\alpha_{\text{vis}} = \alpha_{\text{phy}} = \alpha_{\text{aff}} = \alpha_{\text{cin}} = 0.25$), ensuring a balanced and unbiased assessment across visual appearance, physical causality, affective evolution, and cinematic structure.

%% file: Suppl_secs/extended_exp.tex
\section{Extended Experiments and Analytical Insights}
\label{app:extended_experiments}

To complement the fine-grained per-metric results reported in Table~2 of the main paper, this section provides extended experimental evaluations, dimension-level aggregations, and diagnostic ablations on PersonaShot. Specifically, we present: (1) the complete 12-model leaderboard aggregated across the four specialist dimensions and grouped by prompt conditioning paradigm, (2) fine-grained Emotion-Prompt Alignment (EPA) evaluation using our domain-adapted evaluator, (3) quantitative measurement of within-shot versus cross-shot consistency gaps, and (4) an ablation study on storyboard prompt granularity.

\subsection{Full 12-Model Dimension-Level Leaderboard}
\label{app:full_benchmark}

While Table~2 in the main paper presents detailed sub-metric scores across evaluated systems, Table~\ref{tab:full_leaderboard} summarizes their macro-level performance across the four specialist dimensions: Visual Quality and Consistency (\textbf{D1}), Causal Physical Continuity (\textbf{D2}), Affective Dynamics (\textbf{D3}), and Cinematic Grammar (\textbf{D4}). Following Table~2 in the main paper, models are explicitly categorized by their conditioning paradigm: \textbf{Shot-Level Prompt Conditioning} (storyboard-guided generation) and \textbf{Global-Prompt Conditioning} (single prompt or automated storyboard planning). Strictly following the main paper protocol, the composite score is computed as the unweighted arithmetic mean of the four dimensions ($\alpha_{\text{vis}} = \alpha_{\text{phy}} = \alpha_{\text{aff}} = \alpha_{\text{cin}} = 0.25$).

\begin{table*}[t]
\centering
\small
\caption{\textbf{Complete 12-Model Dimension Leaderboard on PersonaShot.} We report the aggregate Composite score alongside four macro-dimensions across 100 standardized test sequences per model. Models are categorized into \textit{Shot-Level Prompt Conditioning} and \textit{Global-Prompt Conditioning}. Equal dimension weighting ($\alpha = 0.25$) is applied. Within each protocol, the highest score in each column is highlighted in \textbf{bold}.}
\label{tab:full_leaderboard}
\begin{tabular}{c l c c c c c}
\toprule
\textbf{Rank} & \textbf{Model / System} & \textbf{Composite ($\uparrow$)} & \textbf{D1: Visual ($\uparrow$)} & \textbf{D2: Physical ($\uparrow$)} & \textbf{D3: Emotion ($\uparrow$)} & \textbf{D4: Grammar ($\uparrow$)} \\
\midrule
\multicolumn{7}{l}{\textbf{Shot-Level Prompt Conditioning}} \\
\midrule
1  & Seedance (Structured)     & \textbf{0.730} & 0.751 & \textbf{0.824} & \textbf{0.630} & \textbf{0.716} \\
2  & HoloCine (Storyboard)     & 0.707 & \textbf{0.764} & 0.823 & 0.594 & 0.646 \\
3  & STAGE + Wan2.2            & 0.650 & 0.673 & 0.774 & 0.546 & 0.609 \\
4  & LongLive                  & 0.623 & 0.642 & 0.755 & 0.453 & 0.642 \\
5  & MultiShotMaster           & 0.617 & 0.639 & 0.755 & 0.546 & 0.530 \\
6  & STAGE + LTX-2.3           & 0.603 & 0.615 & 0.715 & 0.541 & 0.542 \\
7  & ShotStream                & 0.567 & 0.600 & 0.633 & 0.417 & 0.617 \\
8  & CineTrans                 & 0.554 & 0.602 & 0.637 & 0.429 & 0.550 \\
9  & EchoShot                  & 0.542 & 0.544 & 0.669 & 0.518 & 0.437 \\
\midrule
\multicolumn{7}{l}{\textbf{Global-Prompt Conditioning}} \\
\midrule
1  & LTX-2 (Single Prompt)     & \textbf{0.684} & \textbf{0.722} & 0.786 & 0.482 & \textbf{0.744} \\
2  & Seedance (Single Prompt)  & 0.665 & 0.674 & \textbf{0.808} & \textbf{0.565} & 0.612 \\
3  & VGoT-FP                   & 0.595 & 0.619 & 0.672 & 0.491 & 0.599 \\
\bottomrule
\end{tabular}
\end{table*}

\paragraph{Analytical Insights and Metric Discriminative Power.}
Analyzing the performance differences across conditioning paradigms and dimensions reveals key characteristics of current generators:
\begin{itemize}
    \item \textbf{Structured Prompting vs. Single-Prompt Generation:} Directly supporting Finding 2 of the main paper, models operating under Shot-Level Prompt Conditioning generally outperform their Global-Prompt counterparts in overall narrative consistency. For instance, \textit{Seedance (Structured)} achieves a composite score of \textbf{0.730} compared to \textbf{0.665} for \textit{Seedance (Single Prompt)}, showing significant gains in physical continuity ($0.824 \text{ vs. } 0.808$) and emotional progression ($0.630 \text{ vs. } 0.565$).
    \item \textbf{Cinematic Grammar (D4) exhibits the highest discriminative power:} Among all four dimensions, D4 demonstrates the largest relative variation across models ($\text{CV} = 12.7\%$, ranging from 0.437 to 0.744). Notably, under Global-Prompt Conditioning, \textit{LTX-2 (Single Prompt)} dominates D4 film grammar (\textbf{0.744}) despite lower emotional persistence, indicating strong intrinsic editing priors within open-source base diffusion models.
    \item \textbf{Orthogonality of Narrative Continuity Dimensions:} Pearson correlation analysis indicates that D4 is virtually orthogonal to affective dynamics (\textbf{D3}, $r = -0.07$), confirming that visually and emotionally convincing shots do not automatically form well-sequenced cinematic narratives.
\end{itemize}

\subsection{Fine-Grained Emotion-Prompt Alignment (EPA)}
\label{app:epa_analysis}

To complement general visual-affective scoring, we evaluate Emotion-Prompt Alignment (EPA) using our specialist Qwen3-8B evaluator fine-tuned on the fine-grained MERR dataset (90.6\% validation accuracy across 9 emotional categories). Since EPA evaluation requires explicit per-shot intended emotion descriptions extracted from storyboard captions, Table~\ref{tab:epa_results} reports results across systems operating under \textbf{Shot-Level Prompt Conditioning}.

\begin{table}[ht]
\centering
\small
\caption{\textbf{Emotion-Prompt Alignment (EPA) Results.} We report the composite EPA score alongside exact label match (Hard Acc), valence-arousal similarity (Soft Acc), and emotional arc trajectory correlation (Traj Corr) for shot-level prompt conditioning systems. Best performance is highlighted in \textbf{bold}.}
\label{tab:epa_results}
\resizebox{\linewidth}{!}{%
\begin{tabular}{l c c c c}
\toprule
\textbf{Model / System} & \textbf{EPA ($\uparrow$)} & \textbf{Hard Acc ($\uparrow$)} & \textbf{Soft Acc ($\uparrow$)} & \textbf{Traj Corr ($\uparrow$)} \\
\midrule
Seedance (Structured)  & \textbf{0.802} & \textbf{0.824} & 0.869 & 0.648 \\
STAGE + Wan2.2         & 0.797 & 0.816 & 0.867 & \textbf{0.657} \\
CineTrans              & 0.797 & 0.820 & \textbf{0.870} & 0.646 \\
MultiShotMaster        & 0.796 & 0.815 & 0.864 & 0.651 \\
ShotStream             & 0.796 & 0.814 & 0.863 & 0.651 \\
HoloCine               & 0.793 & 0.815 & 0.864 & 0.637 \\
STAGE + LTX-2.3        & 0.793 & 0.812 & 0.862 & 0.653 \\
LongLive               & 0.788 & 0.804 & 0.856 & 0.653 \\
\bottomrule
\end{tabular}%
}
\end{table}

\paragraph{Analytical Insights.}
As shown in Table~\ref{tab:epa_results}, EPA composite scores cluster tightly across all storyboard-guided systems (ranging from 0.788 to 0.802, a narrow spread of only 0.014). Approximately 81\% to 82\% of generated shots successfully express the intended coarse emotion category (Hard Acc), with soft valence-arousal accuracy exceeding 85\%. However, the trajectory correlation (\text{Traj Corr}) remains moderate ($\sim$0.653) across all models. This confirms that while modern video generators reliably synthesize static emotional expressions within individual shots, orchestrating smooth, narrative-consistent emotional transitions across multi-shot cuts remains a critical bottleneck.

\subsection{Quantitative Gap: Per-Shot vs. Cross-Shot Consistency}
\label{app:temporal_gap}

A primary motivation of PersonaShot is that conventional benchmarks overestimate multi-shot video generation quality by evaluating isolated clips. Table~\ref{tab:temporal_gap} quantifies the performance discrepancy between within-shot (per-shot) metrics and cross-shot relational metrics across the visual and affective dimensions.

\begin{table}[ht]
\centering
\small
\caption{\textbf{Per-Shot vs. Cross-Shot Performance Degradation.} Averaged scores across all evaluated models show substantial drops when transitioning from local shot fidelity to cross-shot relational continuity.}
\label{tab:temporal_gap}
\resizebox{\linewidth}{!}{%
\begin{tabular}{lccc}
\toprule
\textbf{Dimension} & \textbf{Per-Shot Avg ($\uparrow$)} & \textbf{Cross-Shot Avg ($\uparrow$)} & \textbf{Absolute Drop ($\Delta$)} \\
\midrule
D1: Visual Quality        & 0.780 & 0.580 & -0.200 (-25.6\%) \\
D3: Emotional Expression  & 0.700 & 0.350 & \textbf{-0.350 (-50.0\%)} \\
\bottomrule
\end{tabular}%
}
\end{table}

\paragraph{Analytical Insights.}
The data provides direct quantitative confirmation of our main paper findings: across all evaluated systems, cross-shot consistency scores are 25.6\% to 50.0\% lower than per-shot quality scores. While models reliably generate visually appealing individual frames (Per-Shot Visual Avg of 0.780), they frequently reset background styles and character identities across cuts (Cross-Shot Visual Avg of 0.580). This temporal gap is most severe in D3 Emotional Expression, where performance drops by exactly 50.0\% ($0.700 \rightarrow 0.350$). This indicates that micro-expression persistence and emotional arc coherence are the most critical bottlenecks in human-centric multi-shot video generation.

\subsection{Ablation on Storyboard Prompt Granularity}
\label{app:prompt_ablation}

To assess how prompt richness impacts narrative continuity, we conduct an ablation study on the LTX-2 architecture by comparing full storyboard prompts ($\sim$1,200 tokens under Shot-Level Prompt Conditioning) against brief compressed prompts ($\sim$600 tokens retaining only core actions).

\begin{table*}[t]
\centering
\small
\caption{\textbf{Impact of Prompt Length on LTX-2 Performance.} Halving storyboard prompt length causes severe degradation in physical and visual coherence while leaving cinematic grammar relatively resilient.}
\label{tab:prompt_ablation}
\begin{tabular}{l c c c c c}
\toprule
\textbf{Prompt Condition} & \textbf{Composite ($\uparrow$)} & \textbf{D1: Visual ($\uparrow$)} & \textbf{D2: Physical ($\uparrow$)} & \textbf{D3: Emotion ($\uparrow$)} & \textbf{D4: Grammar ($\uparrow$)} \\
\midrule
LTX-2 Full Prompt ($\sim$1200 tokens)  & \textbf{0.705} & \textbf{0.722} & \textbf{0.786} & \textbf{0.482} & \textbf{0.744} \\
LTX-2 Brief Prompt ($\sim$600 tokens) & 0.464          & 0.434          & 0.409          & 0.269          & 0.634          \\
\midrule
\textbf{Absolute Degradation ($\Delta$)} & -0.241 & -0.288 & \textbf{-0.378} & -0.212 & -0.110 \\
\textbf{Relative Performance Drop}     & -34.2\% & -39.9\% & \textbf{-48.0\%} & -44.1\% & -14.8\% \\
\bottomrule
\end{tabular}
\end{table*}

\paragraph{Analytical Insights.}
As reported in Table~\ref{tab:prompt_ablation}, prompt compression degrades overall composite performance by 34.2\%, revealing distinct sensitivities across evaluation dimensions:
\begin{itemize}
    \item \textbf{Physical Consistency is Highly Prompt-Dependent:} D2 suffers the most extreme relative drop (-48.0\%, falling from 0.786 to 0.409). Without explicit storyboard state constraints, objects frequently float, lose physical support, or undergo unexplained shape transformations across cuts.
    \item \textbf{Cinematic Grammar is Robust to Prompt Compression:} In contrast, D4 film grammar exhibits the highest resilience, dropping by only 14.8\% ($0.744 \rightarrow 0.634$). This indicates that camera framing, 180-degree rule compliance, and shot transition logic are largely governed by the model's internal pre-training priors rather than verbose textual instructions.
\end{itemize}

\subsection{Qualitative Visualizations and Cross-Model Comparison}
\label{app:qualitative_figures}

\begin{figure*}[p]
\centering
\includegraphics[width=0.8\textwidth]{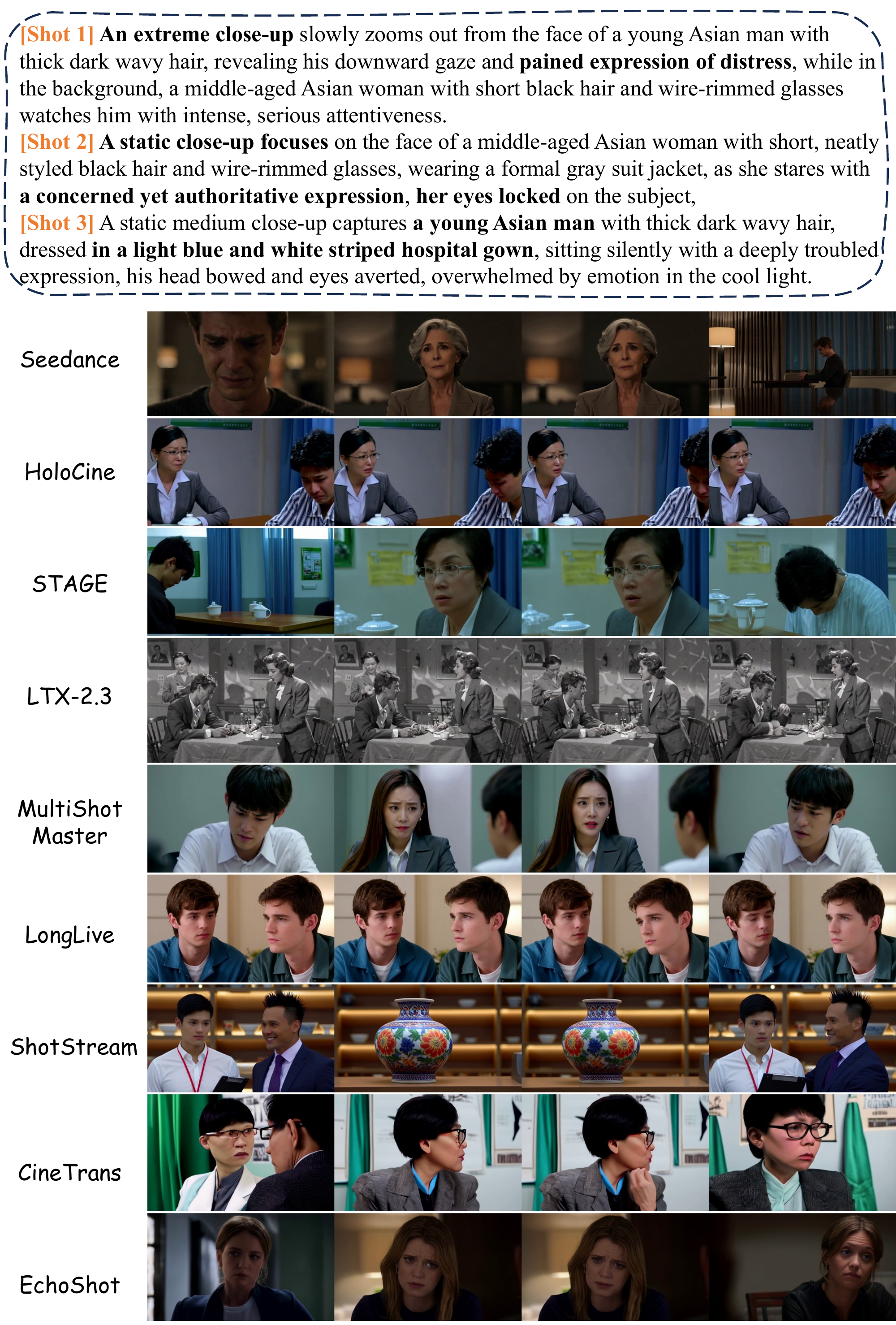}
\caption{\textbf{Qualitative Multi-Model Comparison on a 3-Shot Emotional Dialogue Scene.} Comparison of nine video generation systems given a hierarchical prompt describing an intense emotional interaction between a young man and a middle-aged woman. While \textit{Seedance} and \textit{HoloCine} maintain character identities and emotional distress across cuts, baselines exhibit severe failure modes: \textit{ShotStream} replaces the character with an irrelevant vase, \textit{LTX-2.3} suffers from monochrome style drift, and \textit{EchoShot} experiences sharp identity shifts across shot boundaries.}
\label{fig:qualitative_sample2}
\end{figure*}

\begin{figure*}[p]
\centering
\includegraphics[width=0.8\textwidth]{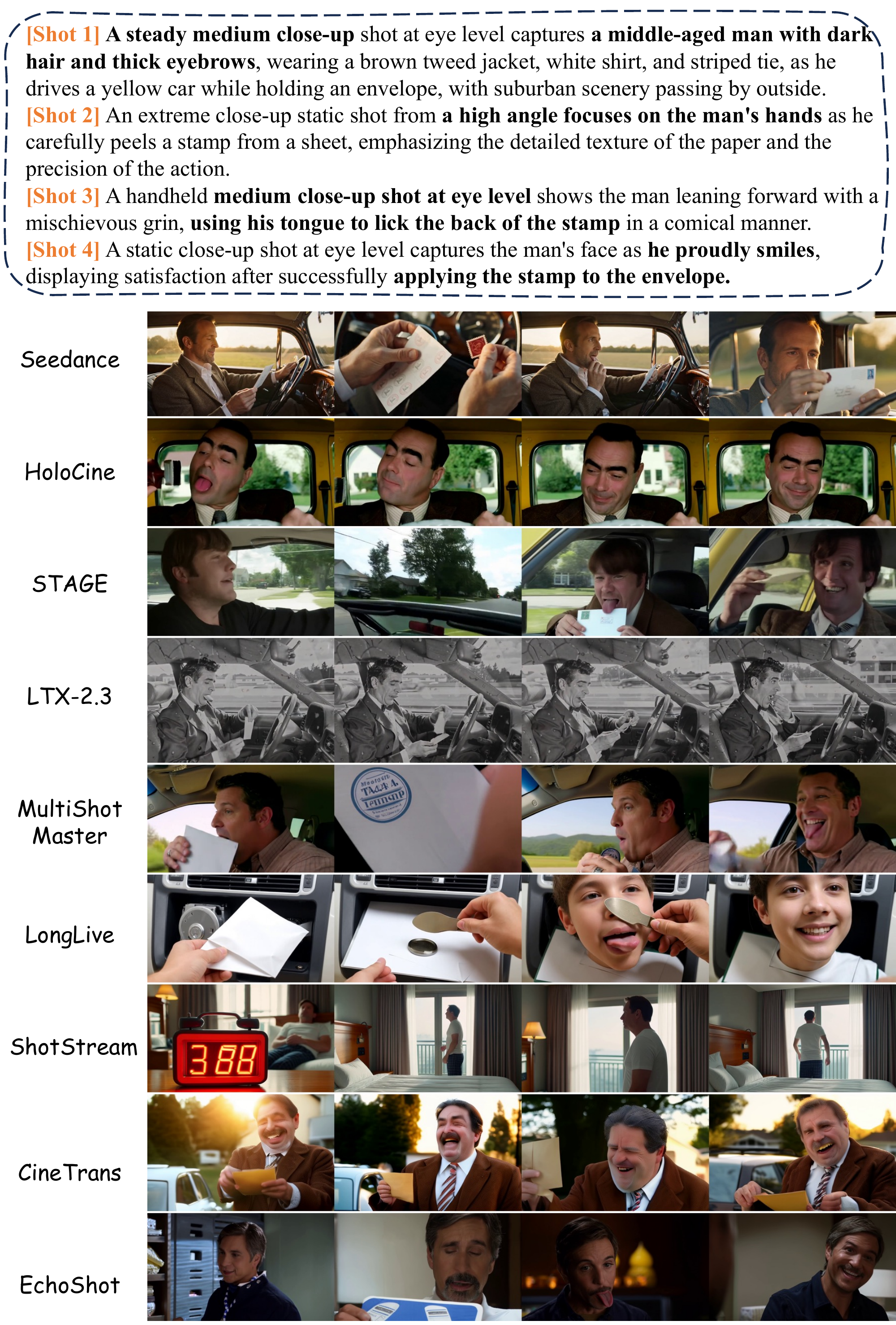}
\caption{\textbf{Qualitative Multi-Model Comparison on a 4-Shot Comical Action Sequence.} Comparison across nine systems generating a multi-shot narrative of a man driving a car, peeling a stamp, licking it, and smiling. \textit{Seedance} and \textit{HoloCine} accurately execute the fine-grained object interaction and facial expressions while preserving character identity. In contrast, \textit{LongLive} hallucinates a child identity in Shot 3, \textit{ShotStream} generates a bedroom alarm clock instead of the car interior, and \textit{STAGE} struggles with stamp-envelope object persistence.}
\label{fig:qualitative_sample1}
\end{figure*}

To provide an intuitive visual validation of our quantitative benchmarks, Figure~\ref{fig:qualitative_sample1} and Figure~\ref{fig:qualitative_sample2} present comprehensive side-by-side qualitative comparisons across nine representative multi-shot video generation systems on multi-shot narrative storyboards.

\paragraph{Qualitative Observations and Failure Diagnostics.}
The visual comparisons corroborate the core quantitative findings established in our benchmark:
\begin{itemize}
    \item \textbf{Superiority of Storyboard-Anchored Consistency:} Systems equipped with structured storyboard guidance, most notably \textit{Seedance} and \textit{HoloCine}, demonstrate remarkable cross-shot narrative continuity. In both dramatic dialogue scenes (Figure~\ref{fig:qualitative_sample2}) and fine-grained action sequences (Figure~\ref{fig:qualitative_sample1}), these models successfully preserve character identity, clothing details, environmental lighting, and emotional progression across cuts.
    \item \textbf{Severe Identity and Age Drift:} Without explicit cross-shot relational reasoning, open-source and streaming baselines frequently suffer from sharp identity resets. For instance, in Figure~\ref{fig:qualitative_sample1}, \textit{LongLive} abruptly shifts the protagonist from a middle-aged man to a young boy in Shots 3 and 4, while \textit{CineTrans} and \textit{EchoShot} exhibit substantial facial and style inconsistencies across shot boundaries.
    \item \textbf{Semantic Hallucination and B-Roll Insertion:} Streaming frameworks such as \textit{ShotStream} struggle to maintain character-centric focus across extended sequences. When prompted with continuous human character actions, \textit{ShotStream} frequently hallucinates disconnected B-roll cutaways, including an irrelevant decorative vase in Figure~\ref{fig:qualitative_sample2} (Shots 2 and 3) or a bedroom digital clock in Figure~\ref{fig:qualitative_sample1} (Shot 1), which breaks character presence entirely.
    \item \textbf{Style and Chromaticity Drift:} While models like \textit{LTX-2.3} capture basic shot sequencing and cinematic framing, they frequently experience chromaticity instability, spontaneously drifting into grayscale or monochrome palettes across cuts (as observed in both Figure~\ref{fig:qualitative_sample1} and Figure~\ref{fig:qualitative_sample2}).
\end{itemize}

\begin{figure*}[t]
\centering
\includegraphics[width=0.85\textwidth]{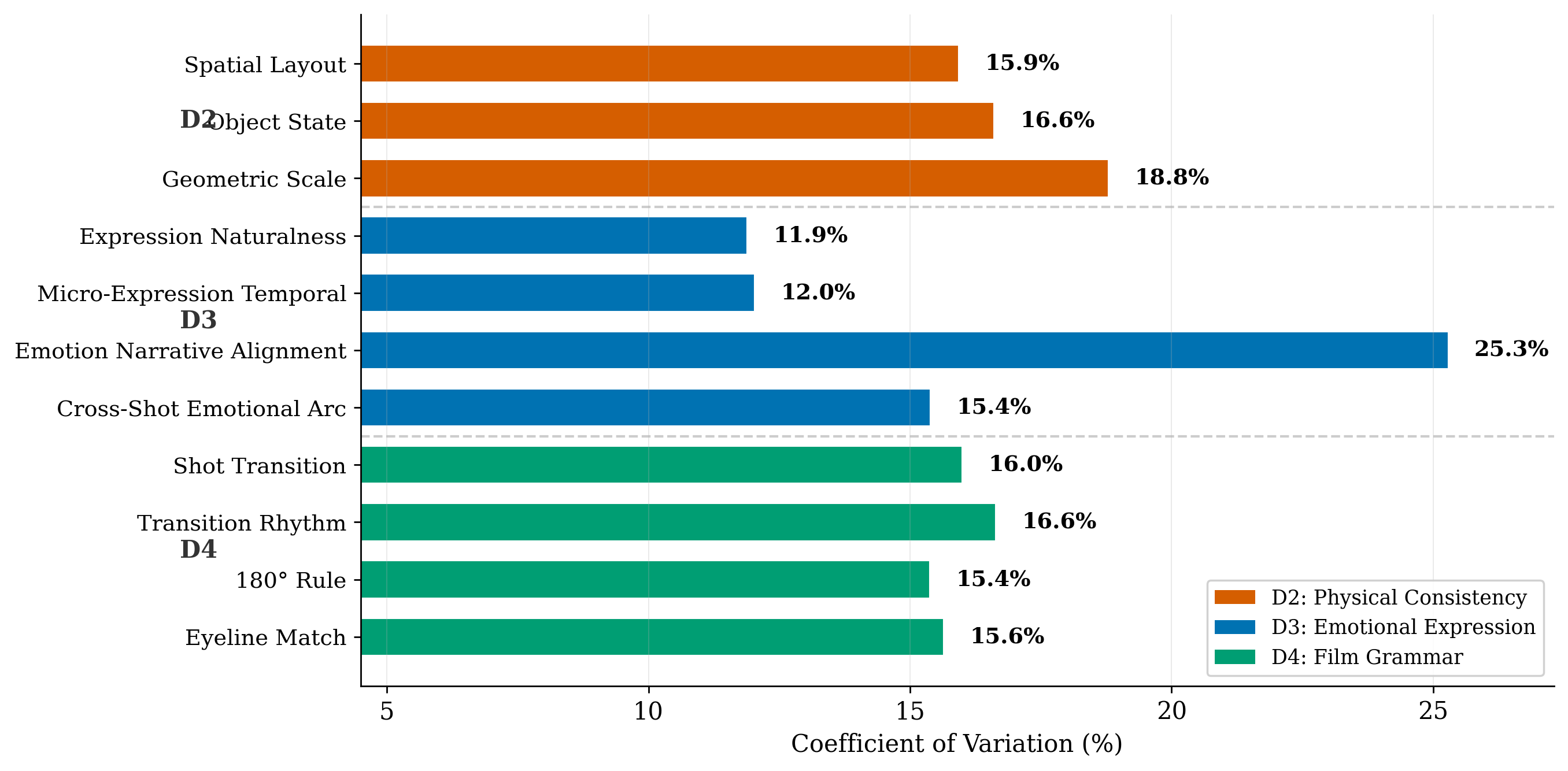}
\caption{\textbf{Coefficient of Variation (CV \%) across PersonaShot Sub-Metrics.} Higher CV indicates stronger discriminative power across evaluated systems. \textit{Emotion Narrative Alignment} ($25.3\%$) exhibits the highest sensitivity across all models.}
\label{fig:cv_barchart}
\end{figure*}

\begin{figure*}[t]
\centering
\includegraphics[width=0.98\textwidth]{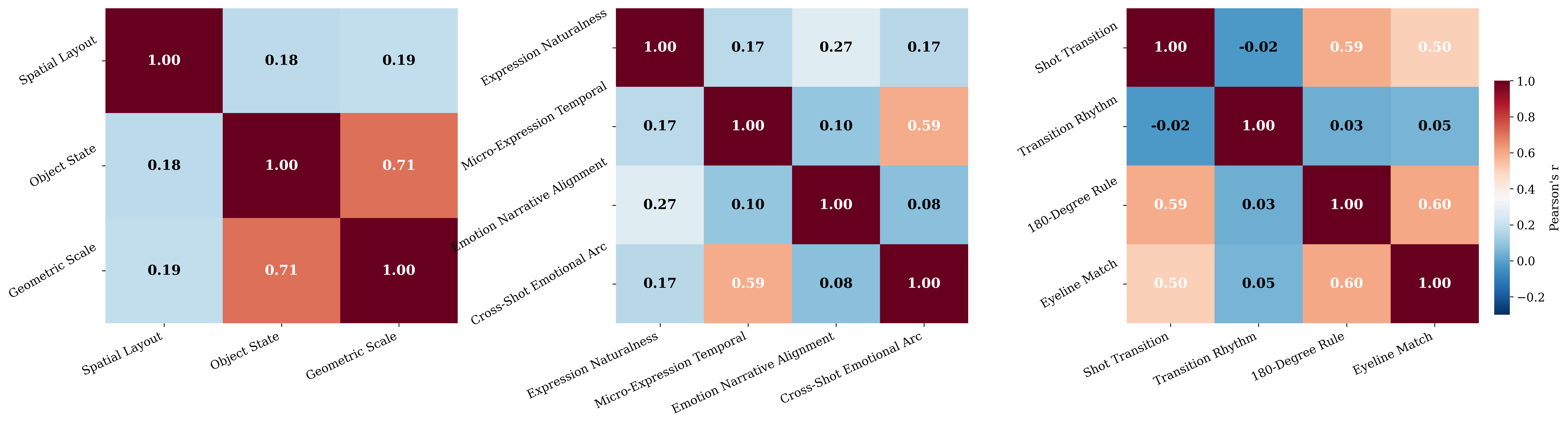}
\caption{\textbf{Internal Correlation Heatmaps (Pearson's $\boldsymbol{r}$) for Specialist Dimensions.} From left to right: Physical Consistency (\textbf{D2}), Emotional Expression (\textbf{D3}), and Film Grammar (\textbf{D4}). Low correlations across multiple sub-metrics confirm the orthogonality and non-redundancy of our benchmark criteria.}
\label{fig:internal_correlation}
\end{figure*}

\subsection{Metric Sensitivity and Internal Orthogonality Analysis}
\label{app:metric_sensitivity_orthogonality}

To verify the discriminative power and structural non-redundancy of our evaluation framework, we analyze the Coefficient of Variation (CV) across all sub-metrics (Figure~\ref{fig:cv_barchart}) alongside internal pairwise correlations within each specialist dimension (Figure~\ref{fig:internal_correlation}).

\paragraph{Metric Discriminative Power (Coefficient of Variation).}
As shown in Figure~\ref{fig:cv_barchart}, the CV values reveal significant differences in metric sensitivity across evaluated systems:
\begin{itemize}
    \item \textbf{High Variance in Narrative Alignment:} \textit{Emotion Narrative Alignment (ENA)} exhibits the highest CV across the benchmark ($25.3\%$). While most models achieve comparable baseline performance in static facial realism (\textit{Expression Naturalness}, $\text{CV} = 11.9\%$) and local stability (\textit{Micro-Expression Temporal}, $\text{CV} = 12.0\%$), their capabilities diverge sharply when tasked with matching emotional category and intensity to complex narrative contexts.
    \item \textbf{Sensitivity of Geometric Scaling:} Within Physical Consistency (\textbf{D2}), \textit{Geometric Scale} exhibits the highest variation ($18.8\%$), reflecting severe scale drift and character-to-object proportion distortion across viewpoint changes in weaker models.
    \item \textbf{Uniform Difficulty of Cinematic Grammar:} Sub-metrics within Film Grammar (\textbf{D4}) maintain balanced CV values between $15.4\%$ and $16.6\%$, indicating that temporal cutting rhythms and spatial camera rules represent uniformly challenging bottlenecks for current video generators.
\end{itemize}

\paragraph{Internal Orthogonality and Synergy (Pearson's $\boldsymbol{r}$).}
The correlation heatmaps in Figure~\ref{fig:internal_correlation} demonstrate that our sub-metrics evaluate complementary rather than redundant capabilities:
\begin{itemize}
    \item \textbf{Physical Consistency (D2):} \textit{Spatial Layout} is nearly orthogonal to \textit{Object State} ($r = 0.18$) and \textit{Geometric Scale} ($r = 0.19$). This confirms that preserving correct character and object positions across cuts does not guarantee morphological stability. Conversely, \textit{Object State} and \textit{Geometric Scale} exhibit a strong positive correlation ($r = 0.71$), as shape deformation and state resets frequently occur alongside relative scale distortions.
    \item \textbf{Emotional Expression (D3):} \textit{Emotion Narrative Alignment} shows low correlation with low-level facial execution metrics ($r \le 0.27$), proving that semantic emotion-to-story adherence is independent of visual expression rendering. Meanwhile, \textit{Micro-Expression Temporal} and \textit{Cross-Shot Emotional Arc} correlate moderately ($r = 0.59$), reflecting shared requirements for short- and long-term temporal continuity.
    \item \textbf{Film Grammar (D4):} \textit{Transition Rhythm} is completely decoupled from spatial cinematic rules ($r \in [-0.02, 0.05]$). This demonstrates that temporal cutting pacing (rhythm) and spatial camera geometry (axis and eyeline matching) assess fundamentally distinct directorial skills. In contrast, \textit{Shot Transition}, \textit{180-Degree Rule}, and \textit{Eyeline Match} share moderate positive correlations ($r \in [0.50, 0.60]$), as they jointly govern cross-shot spatial coherence.
\end{itemize}

%% file: Suppl_secs/human_study.tex
\section{Additional Human Study Details}
\label{app:supp_human}

To verify that our specialist evaluators accurately reflect human judgment, we conduct an expert evaluation study on a held-out set of 25 multi-shot video sequences. This section outlines the evaluation protocol and presents both dimension-level and fine-grained criterion-level alignment between our automatic scores and human preference.

\subsection{Evaluation Protocol}
Ten domain experts participated in the study. Under a sparse assignment protocol, each video sequence was independently rated by four annotators across 12 specialist criteria using a five-point Likert scale. Model identities and automatic scores were strictly concealed.

To account for criteria that require specific visual or relational context, annotators could mark a metric as Not Applicable (N/A) if the corresponding evidence was absent. Such instances were dynamically excluded from correlation calculations rather than assigned default values. Across all evaluated sequences, the pooled inter-rater reliability reached Krippendorff's $\alpha = 0.76$, indicating high consensus among experts.

\subsection{Alignment with Human Judgments}
We evaluate alignment using Spearman rank correlation ($\rho$) and pairwise preference accuracy between automatic scores and expert Mean Opinion Scores (MOS). Table~\ref{tab:human_dimension_alignment} reports the aggregated dimension-level results, while Table~\ref{tab:human_fine_grained_alignment} details the performance across all 12 fine-grained criteria.

\begin{table}[h]
\centering
\caption{\textbf{Dimension-level alignment with human MOS.} All correlations are statistically significant at $p < 0.01$.}
\label{tab:human_dimension_alignment}
\small
\begin{tabular}{lccc}
\toprule
\textbf{Dimension} & $\boldsymbol{N}$ & $\boldsymbol{\rho}\uparrow$ & \textbf{Pair Acc.}$\uparrow$ \\
\midrule
Causal Physical Continuity & 25 & 0.75 & 75.2\% \\
Affective Dynamics         & 25 & 0.69 & 70.3\% \\
Cinematic Grammar          & 25 & 0.73 & 72.8\% \\
\midrule
\textbf{Overall PersonaShot} & \textbf{25} & \textbf{0.78} & \textbf{76.8\%} \\
\bottomrule
\end{tabular}
\end{table}

\begin{table}[h]
\centering
\caption{\textbf{Fine-grained criterion-level human alignment.} Comparison of automatic evaluator scores against expert MOS across all 12 specialist criteria. Correlations are computed over valid evidence-gated sequences ($p < 0.01$).}
\label{tab:human_fine_grained_alignment}
\small
\setlength{\tabcolsep}{8pt}
\begin{tabular}{llcc}
\toprule
\textbf{Dim.} & \textbf{Criterion} & $\boldsymbol{\rho}\uparrow$ & \textbf{Pair Acc.}$\uparrow$ \\
\midrule
\multirow{3}{*}{\textit{Phy.}} 
 & Spatial Layout (SLC)      & 0.68 & 71.4\% \\
 & Object State (OSP)        & 0.74 & 76.0\% \\
 & Geometric Scale (GSC)     & 0.66 & 69.8\% \\
\midrule
\multirow{4}{*}{\textit{Aff.}} 
 & Expr. Naturalness (EN)    & 0.70 & 71.8\% \\
 & Action Coherence (FATC)   & 0.68 & 64.8\% \\
 & Narrative Align. (ENA)    & 0.71 & 72.5\% \\
 & Arc Coherence (EAC)       & 0.67 & 68.2\% \\
\midrule
\multirow{5}{*}{\textit{Cin.}} 
 & Narrative Seq. (DNS)      & 0.74 & 74.0\% \\
 & Shot Transition (ST)      & 0.72 & 73.1\% \\
 & Rhythm Align. (TRA)       & 0.69 & 70.5\% \\
 & 180-Degree Rule (180R)    & 0.71 & 71.9\% \\
 & Eyeline Match (EM)        & 0.65 & 67.2\% \\
\bottomrule
\end{tabular}
\end{table}

\paragraph{Analytical Insights.}
The results across both tables highlight several key characteristics of our evaluation framework:
\begin{itemize}
    \item \textbf{Strong Holistic Agreement:} As shown in Table~\ref{tab:human_dimension_alignment}, the overall PersonaShot score achieves a Spearman correlation of 0.78 and a pairwise accuracy of 76.8\%. This confirms that combining our specialized dimensions effectively reflects holistic human preference in multi-shot narrative continuity.
    \item \textbf{High Precision on Explicit Rules:} Table~\ref{tab:human_fine_grained_alignment} demonstrates that evaluators grounded in explicit structural logic achieve remarkably strong alignment with human experts. In particular, Object State (OSP, $\rho = 0.74$), Directorial Narrative Sequencing (DNS, $\rho = 0.74$), and Shot Transition (ST, $\rho = 0.72$) show high pairwise accuracy above 73\%, proving that our instruction-tuned reasoners successfully capture temporal state evolution and editing syntax.
    \item \textbf{Validity of Evidence-Gating:} Crucially, correlations are computed strictly over applicable, evidence-gated sequences (e.g., evaluating 180-Degree Rule and Eyeline Match only on the subset of sequences containing shot-reverse-shot interactions). Masking out inapplicable instances prevents random scoring noise and ensures reliable alignment.
\end{itemize}